\documentclass[sigplan,10pt]{acmart}

\AtBeginDocument{%
  }

\usepackage{tikz}
\usepackage{amsmath}
\usepackage{enumitem}
\usepackage{filecontents}
\usepackage{xspace}
\usepackage{float}
\usepackage{booktabs}
\usepackage{graphicx}
\usepackage{subcaption}
\usepackage{amsmath}
\usepackage{balance}

\usepackage{amssymb}
\usepackage{makecell}
\usepackage{array}
\usepackage{listings}
\usepackage[ruled,linesnumbered]{algorithm2e}
\newcommand{\comments}[2]{{\bf[\textcolor{red}{#1}: \textcolor{blue}{#2}]}}

\newcommand{\hao}[1]{{\comments{Hao}{#1}}}

\newcommand{\system}{STAlloc\xspace}
\newcommand{\para}[1]{{\vspace{5pt} \bf \noindent #1 \hspace{5pt}}}
\newcommand{\fixme}[1]{{\color{red} #1}}
\newcommand{\squishlist}{
	\begin{list}{$\bullet$}
		{ \setlength{\itemsep}{0pt}      \setlength{\parsep}{3pt}
			\setlength{\topsep}{3pt}       \setlength{\partopsep}{0pt}
			\setlength{\leftmargin}{3.5mm} \setlength{\labelwidth}{1em}
			\setlength{\labelsep}{0.5em} } }
	\newcommand{\squishend}{
\end{list}  }
\acmYear{2026}
\copyrightyear{2026}
\setcopyright{cc}
\setcctype{by}
\acmConference[EUROSYS '26]{European Conference on Computer Systems}{April 27--30, 2026}{Edinburgh, Scotland Uk}
\acmBooktitle{European Conference on Computer Systems (EUROSYS '26), April 27--30, 2026, Edinburgh, Scotland Uk}
\acmDOI{10.1145/3767295.3769335}
\acmISBN{979-8-4007-2212-7/26/04}
\acmSubmissionID{192}
\begin{document}


% Origin Title
\title{\system: Enhancing Memory Efficiency in Large-Scale Model Training with Spatio-Temporal Planning}

% Arxiv Title
% \title{Reducing GPU Memory Fragmentation via Spatio-Temporal Planning for Efficient Large-Scale Model Training}

\author{Zixiao Huang}
\authornote{Both authors contributed equally to this research.}
\email{huangzx21@mails.tsinghua.edu.cn}
\affiliation{%
  \institution{Tsinghua University}
  \institution{Infinigence AI}
  % \city{Beijing}
  \country{ }
}

\author{Junhao Hu}
\authornotemark[1]
\email{hujunhao@infini-ai.com}
\affiliation{%
  \institution{Infinigence AI}
  % \city{Beijing}
  \country{ }
}

\author{Hao Lin}
\email{linhao@infini-ai.com}
\affiliation{%
  \institution{Tsinghua University}
  \institution{Infinigence AI}
  % \city{Rocquencourt}
  \country{ }
}

\author{Chunyang Zhu}
\email{zhuchunyang@infini-ai.com}
\affiliation{%
 \institution{Infinigence AI}
 \country{ }
 }

\author{Yueran Tang}
\email{tangyueran@infini-ai.com}
\affiliation{%
  \institution{Infinigence AI}
  \country{ }
}

\author{Quanlu Zhang}
\email{zhangquanlu@infini-ai.com}
\authornote{Corresponding authors.}
\affiliation{%
  \institution{Infinigence AI}
  \country{ }
}

\author{Zhen Guo}
\email{guozhen@infini-ai.com}
\affiliation{%
  \institution{Infinigence AI}
  \country{ }
}

\author{Zhenhua Li}
\email{lizhenhua1983@tsinghua.edu.cn}
\affiliation{%
  \institution{Tsinghua University}
  \country{ }
}

\author{Shengen Yan}
\email{yansg@mail.tsinghua.edu.cn}
\affiliation{%
  \institution{Tsinghua University}
  \institution{Infinigence AI}
  \country{ }
}

\author{Zhenhua Zhu}
\email{zhuzhenhua@mail.tsinghua.edu.cn}
\authornotemark[2]
\affiliation{%
  \institution{Tsinghua University}
  \country{ }
}

\author{Guohao Dai}
\email{daiguohao@sjtu.edu.cn}
\affiliation{%
  \institution{Shanghai Jiao Tong University}
  \institution{Infinigence AI}
  \country{ }
}

\author{Yu Wang}
\email{yu-wang@mail.tsinghua.edu.cn}
\authornotemark[2]
\affiliation{%
  \institution{Tsinghua University}
  \country{ }
}

\renewcommand{\shortauthors}{Zixiao Huang, Junhao Hu, Hao Lin, Chunyang Zhu, Yueran Tang, Quanlu Zhang et al.}
\renewcommand{\shorttitle}{\system}

\begin{CCSXML}
<ccs2012>
   <concept>
       <concept_id>10011007.10010940.10010941.10010949.10010950</concept_id>
       <concept_desc>Software and its engineering~Memory management</concept_desc>
       <concept_significance>500</concept_significance>
       </concept>
   <concept>
       <concept_id>10010520.10010521.10010528</concept_id>
       <concept_desc>Computer systems organization~Parallel architectures</concept_desc>
       <concept_significance>300</concept_significance>
       </concept>
   <concept>
       <concept_id>10010147.10010257</concept_id>
       <concept_desc>Computing methodologies~Machine learning</concept_desc>
       <concept_significance>100</concept_significance>
       </concept>
 </ccs2012>
\end{CCSXML}

\ccsdesc[500]{Software and its engineering~Memory management}
\ccsdesc[300]{Computer systems organization~Parallel architectures}
\ccsdesc[100]{Computing methodologies~Machine learning}


\keywords{Memory Defragmentation, Large Language Model, GPU, Distributed Training}


\begin{abstract}
The rapid scaling of large language models (LLMs) has significantly increased GPU memory pressure, 
    which is further aggravated by training optimization techniques such as virtual pipeline and recomputation that disrupt tensor lifespans and introduce considerable memory fragmentation. 
Such fragmentation stems from the use of online GPU memory allocators in popular deep learning frameworks like PyTorch, which disregard tensor lifespans. 
As a result, this inefficiency can waste as much as 43\% of memory and trigger out-of-memory errors, undermining the effectiveness of optimization methods.
    
To address this, we introduce \system, a GPU memory allocator for deep learning frameworks that reduces fragmentation by exploiting the spatial and temporal regularity in memory allocation behaviors of training workloads. 
% Unlike existing memory allocators that rely entirely on online allocation strategies, 
\system introduces a novel paradigm that combines offline planning with online allocation. 
The offline planning leverages spatio-temporal regularities to generate a near-optimal allocation plan, while the online allocation handles complex and dynamic models such as Mixture-of-Experts (MoE).
% \system plans allocations for static memory requests (recurrent same-size tensors with predictable lifespans) using a grouping algorithm that overlaps non-conflicting temporal intervals, and dynamically handles irregular requests with a lifespan-aware strategy. 
% For dynamic models such as the mixture-of-experts, \system safely overlaps memory regions by leveraging temporal regularity. 
Built as a pluggable PyTorch memory allocator, \system reduces fragmentation ratio on average by 85.1\% (up to 100\%) across both dense and MoE models, with negligible overhead. 
This enables more efficient, high-throughput training configurations and improves throughput performance by up to 32.5\%.

\end{abstract}

\maketitle

%-------------------------------------------------------------------------------
\begin{figure}[t]
    \begin{subfigure}[b]{0.45\linewidth}
        \centering
        \includegraphics[width=\linewidth]{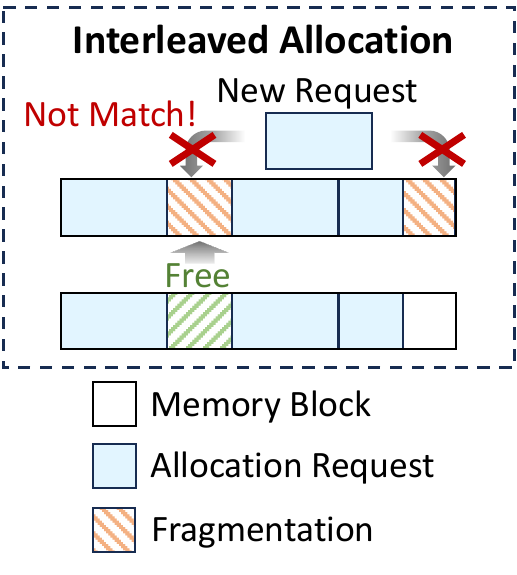}
        %\caption{Memory fragmentation due to memory lifecycle-disrupting optimizations in the caching allocator.}
        \label{fig:mem_frag_example}
        % \vspace{-15pt}

    \end{subfigure}
    \begin{subfigure}[b]{0.5\linewidth}
        \includegraphics[width=\linewidth]{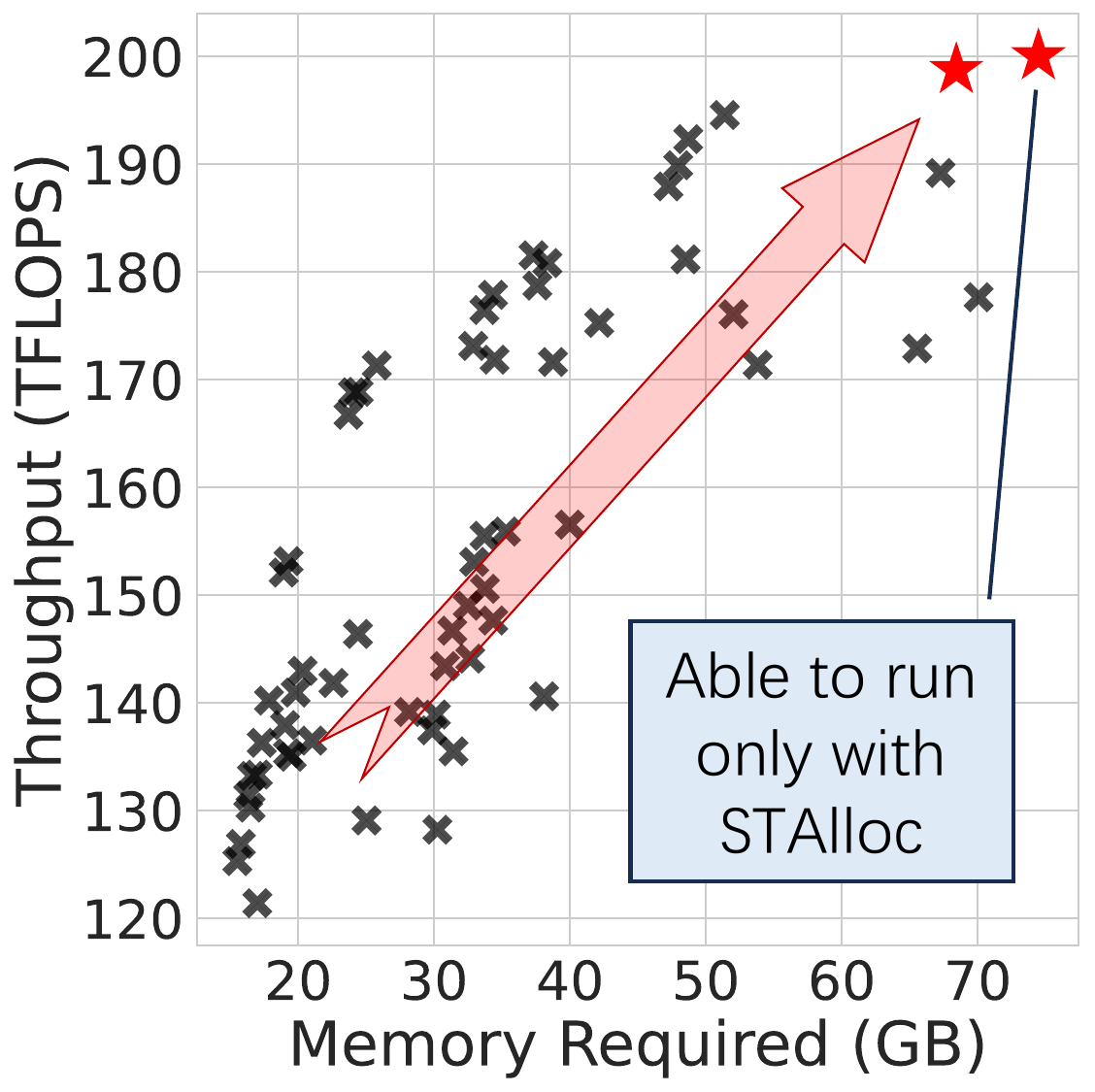}
        %\caption{Theoretical memory consumption and training throughput of different training configurations for Llama2-7B on 8 NVIDIA A800 GPUs.}
        \label{fig:memory&throughput}
        % \vspace{-15pt}

    \end{subfigure}
    \vspace{10pt}
    \caption{(a) Memory fragmentation in interleaved allocation. (b) Memory and training throughput of different training configurations for Llama2-7B on 8 NVIDIA A800 GPUs.}
    % \vspace{-10pt}
    \label{fig:fig1}
\end{figure}

\section{Introduction}
\label{sec:intro}

In recent years, large-scale models, particularly large language models (LLMs)~\cite{llama,gpt4,mistral,gemini,gemma,qwen,deepseek-v3,glm},
    have demonstrated extraordinary performance in language comprehension, problem reasoning, code generation, etc.
The scaling law~\cite{scaling-law} dictates that such powerful capabilities stem from the models' massive parameters and training data.
As a result, nowadays even a medium-sized model such as Llama-3~\cite{llama3} with 70 billion parameters requires more than 1~TB GPU/accelerator memory for training, placing heavy demands on the scarce and expensive GPU memory resource.

Additionally, current large-scale model training often employs a combination of various optimization techniques to enhance overall training efficiency. 
Such optimization techniques serve to either boost training throughput~\cite{megatron2, zero-bubble-pipeline} or reduce the theoretical GPU memory demand of the training~\cite{recompute1, offload1, offload2, zero, megatron3}. 
For instance, the Virtual Pipeline~\cite{megatron2} partitions a conventional pipeline parallel stage into several virtual stages, thereby minimizing idle periods (i.e., pipeline bubbles) inherent in pipeline parallelism. 
Furthermore, memory optimization techniques such as recomputation~\cite{recompute1}, tensor offloading~\cite{offload1, offload2}, and ZeRO~\cite{zero} trade additional computation or transmission for reduced GPU memory usage.

However, the application of these training optimization techniques alters GPU memory allocation patterns. First, the number of allocation requests increases significantly compared to the training configuration without these techniques (e.g., 30\% increase). Second, the allocation pattern shifts from a regular sequence of allocations followed by deallocations (e.g., activation tensors reserved for backward computation) to a more complex, interleaved pattern with frequent alternation between the two.

Unfortunately, the memory allocators in current deep learning frameworks, such as PyTorch~\cite{pytorch}, struggle to efficiently handle such complex allocation patterns, leading to severe memory fragmentation (up to 43\% in typical scenarios). 
Consequently, the actual memory consumption during training significantly exceeds the theoretical allocation requirements.
The root cause of fragmentation lies in the online best-fit allocation policy adopted by the allocator in popular deep learning frameworks (e.g., PyTorch).
This policy allocates a requested tensor of a certain size to the most suitable memory slot without considering the tensor's lifespan, which is unknown to the allocator.
Unpredictable deallocations lead to a discontinuous memory space, making it difficult to fit new tensors, as illustrated in Figure~\ref{fig:fig1}(a).
Over time, this increases fragmentation as free space becomes scattered and less reusable for larger requests.

More critically, the increased GPU memory consumption caused by fragmentation can slow down model training. In large-scale training, configurations with higher throughput often require more GPU memory, as shown in Figure~\ref{fig:fig1}(b), where each point represents a different setup, i.e., using different optimization techniques. Fragmentation reduces the amount of available GPU memory, limiting the feasibility of high-throughput configurations. When such configurations are used, fragmentation can cause actual memory usage to far exceed theoretical estimates, leading to out-of-memory (OOM) errors. As a result, model developers are forced to revert to less efficient configurations with extra computation or communication, thus reducing training throughput (e.g., up to 24.5\%).

To address these problems, we propose \system, a novel GPU memory allocator for deep learning frameworks to reduce fragmentation.
Our approach is based on the observation that GPU memory requests exhibit strong consistency across training iterations. 
Therefore, by pre-assigning memory addresses before training, we can reduce fragmentation caused by online allocation in current allocators. 

However, optimizing memory allocation requests ahead of training meets two challenges. First, offline allocation planning is NP-hard, known as Dynamic Storage Allocation problem~\cite{DSA-problem}. 
In large-scale model training, the number of memory requests can exceed $10^5$, making direct optimization intractable.
To obtain a near-optimal solution within an acceptable time, we extract spatio-temporal regularities from memory allocation during training and use them to guide a grouping-based optimization. 
This grouping approach decomposes the time and space characteristics of memory requests, significantly reducing the complexity of the optimization problem.

Second, the recent emergence of sparse models of Mixture-of-Experts (MoE) models~\cite{mistral,deepseek-v3,deepseek-v2} introduces dynamics in memory allocation patterns compared to dense models.
MoE models replace MLP layers with expert layers,
    and decide which experts to use for each token at runtime, which results in the dynamic nature of allocation request sizes.
Consequently, we cannot rely on planning of certain address for the allocation requests.
To address the challenge of dynamic request sizes, we propose a hybrid paradigm that combines offline planning with online allocation. 
By identifying reusable regions for dynamic requests before training and performing online allocation at runtime, \system supports the dynamicity of allocation requests while maintaining a low fragmentation rate.

We implement \system as a pluggable memory allocator for PyTorch and evaluate it across over 50 training configurations on three different testbeds. 
These configurations combine diverse dense and MoE models, model sizes, optimization techniques, microbatch sizes, and training frameworks. 
\system reduces fragmentation memory by an average of 85.1\% (up to 100\%), saving up to 56.3GB GPU memory with negligible impact on end-to-end training throughput. 
By reducing peak GPU memory usage, it enables efficient training configurations that would otherwise trigger Out-of-Memory errors, resulting in an up to 32.5\% throughput improvement. 
We open source \system to support more developers' efficient large-scale training\footnote{https://github.com/infinigence/STAlloc}.

This paper makes three main contributions:
\squishlist
\item We conduct an in-depth analysis of the memory allocation characteristics and fragmentation problem of large model training,
    identifying spatial and temporal regularity in the allocation pattern.
% \item We design a novel memory allocation algorithm for large-scale model training, which can deal with both static and dynamic memory allocation patterns and greatly reduce fragmentation.
\item We propose a memory allocation paradigm for large-scale model training that combines offline planning with online allocation. 
\system is capable of generating a near-optimal allocation plan  based on spatio-temporal regularities, while effectively accommodating the dynamicity of allocation requests at runtime.
\item We comprehensively evaluated \system using diverse training configurations on different testbeds, demonstrating its wide applicability and effectiveness. It also enables more efficient model training.
\squishend

%-------------------------------------------------------------------------------
% \vspace{-10pt}
\section{Background and Motivation}

\subsection{Memory-driven Parallelism and Optimization}
The evolution of distributed training parallelism has been driven by the critical need to fit increasingly large models into limited GPU memory. 
Early data parallelism (DP) strategies, which replicate the entire model, became infeasible for large-scale training. 
This led to model parallelism techniques such as tensor parallelism (TP)~\cite{megatron}, which partitions weights, and pipeline parallelism (PP)~\cite{pipedream-1f1b, gpipe, zero-bubble-pipeline}, which distributes layers, each with distinct memory trade-offs. 
Optimization techniques like virtual pipeline parallelism (VPP)~\cite{megatron2} further optimize pipeline scheduling to reduce bubbles and improve throughput, though the more complex scheduling often increases memory usage. 
To address the escalating memory demands from models like Mixture-of-Experts (MoE) or those with long sequences, more advanced methods emerged. 
These include expert parallelism (EP)~\cite{gshard} for distributing experts, sequence parallelism (SP)~\cite{megatron3} for sharding activations, and ZeRO~\cite{zero} optimizations, which partition optimizer states, gradients, and even weights. 
This progression clearly demonstrates that efficient GPU memory usage is the central consideration shaping the design of modern parallel training systems.

While parallelism strategies help distribute memory load across GPUs, the number of available GPUs is often limited. To make training fit within GPU memory, recomputation~\cite{recompute1} and tensor offloading~\cite{offload1, offload2, offload3} are commonly used to reduce GPU memory usage, at the cost of slower training. 
Recomputation involves recalculating activation tensors within model layers during backpropagation rather than storing them, allowing for memory savings.
The tensor offloading technique temporarily shifts tensors to CPU memory and retrieves them back when needed. %Unfortunately, even with careful combinations of parallelism, recomputation, and offloading, achieving the desired balance between memory efficiency and training speed often remains elusive, as discussed below.
Unfortunately, even with careful and reasonable combinations of parallelism, recomputation, and offloading, the desired training configuration often encounters the OOM error due to less effective usage of GPU memory, thus falling back to a less efficient training configuration.

\subsection{Low Memory Efficiency in LLM Training}
\label{sec:low-mem-eff}
\begin{figure}[t]
    \centering
    \includegraphics[width=0.7\linewidth]{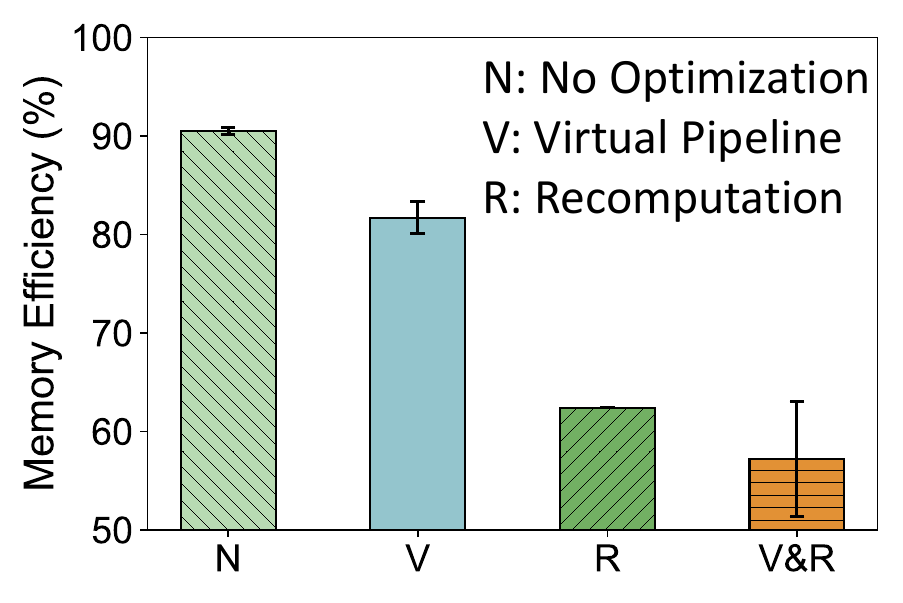}
    \vspace{-10pt}
    \caption{Comparison of PyTorch memory efficiency with no optimizations, recomputation, and Virtual Pipeline.}
    \vspace{-10pt}
    \label{fig:mem_util_comp}
\end{figure}

When training large models on GPUs, operators generate tensors of varying sizes and lifespans, which must be managed in GPU memory. 
These allocation requests pose significant challenges for memory allocators of current deep learning frameworks (e.g., PyTorch). 
Lacking prior knowledge of allocation patterns, allocators typically use online allocation strategy~\cite{pytorch}. 
To reduce system call overhead (e.g., \texttt{cudaMalloc}), they often pre-allocate large caching blocks and slice out chunks based on best fit policy~\cite{best-fit-allocation}. 
Over time, this results in memory fragmentation, where free regions become too small or scattered to satisfy new allocation requests.
For clarity, we define \emph{memory efficiency} ($E$) as the ratio of the actual allocated tensor size to the reserved GPU memory size, which is: 
\begin{align}
    E = \frac{M_a}{M_r}
\end{align}
where $M_a$ is the size of allocated memory, representing the theoretical memory required under current training configuration; $M_r$ is the total memory reserved by the allocator, representing the actual memory usage.

Large model training often leads to severe GPU memory fragmentation, especially when complex parallelism strategies are combined with memory optimization techniques (e.g., recomputation). 
Figure~\ref{fig:mem_util_comp} shows the memory efficiency of GPT-2 (345M parameters)~\cite{openai-gpt2} trained on 8 NVIDIA A800-80G GPUs under different training configurations. 
The baseline uses 1F1B pipeline parallelism, achieves acceptable 90\% memory efficiency with 52.1~GB of reserved memory.
Using Virtual Pipeline Parallelism (VPP)~\cite{megatron2} can improve training throughput.
However, the utilization of VPP increases the allocated memory to 51.8~GB, and complicates the memory activities, reducing memory efficiency to 80\%, leading to 59.9 GB of reserved memory.
This higher memory usage can lead to OOM errors in some training scenarios. 
Recomputation is often used to mitigate memory requirement; however, while it reduces allocated memory, it also drops memory efficiency around 60\%, causing significant memory waste. 
Therefore, memory fragmentation prevents logically effective approaches from achieving the expected memory reduction.
Not only GPT-2, we found that many popular large models (e.g., Llama~\cite{llama}, Qwen~\cite{qwen}) suffer from serious memory fragmentation in training (see \S~\ref{sec:evaluation}).

\noindent
\textbf{Low Memory Efficiency Slows Training.}
Low memory efficiency often prevents more efficient parallelism strategies from fitting within available GPUs, which is a common challenge in large model training. 
As a case, we trained Qwen2.5-14B on 16 NVIDIA H200 GPUs, requiring at least 2-way tensor parallelism ($TP=2$) to fit. 
We selected 2-way pipeline parallelism ($PP=2$), 4-way data parallelism ($DP=4$), and enabled VPP to reduce bubble ratio for better training speed, but encountered OOM. 
To adapt, we tried three alternatives: (1) replacing VPP with 1F1B, but still occurs OOM, (2) enabling recomputation, and (3) increasing $TP$ from 2 to 4. 
% We also applied \system to $TP=2$, $PP=2$, and $DP=4$ with virtual pipeline enabled, which successfully fit in memory. 
The alternative training configurations degrade training speed by 24.5\% and 7.1\% compared to the ideal case without memory fragmentation, respectively, highlighting the critical role of memory efficiency in enabling high-performance parallelism strategies.

% \para{Dynamic allocation leads to unpredictable OOM.}
% The allocation behavior of online allocators, such as caching allocators, is inherently non-deterministic across training iterations, as the (de)allocations of each iteration influence subsequent ones. As shown in Figure 3(a), fragmentation typically accumulates over the first several iterations before stabilizing. This often leads to a counterintuitive scenario, where early iterations succeed but a later iteration fails with an out-of-memory (OOM) error, as illustrated in Figure 3(b). In principle, if the first iteration fits in memory, subsequent iterations should also fit, highlighting the need for a more efficient memory allocator.
% The allocation behaviors of online allocators, such as caching allocator, are non-deterministic for each training iteration. The (de)allocation behaviors of the former iterations affect the behaviors of the latter iterations. Usually, fragmentation granually increases in the first several iterations then becomes stable in the following iterations, as shown in Figure 3(a). Therefore, it is a common phenomenon that the training can fit in the first several iterations while encounters out-of-memory error in the certain iteration behind. Figure 3(b) shows an example of this case. Theoretically if the first iteration can fit, the following iterations should be able to fit too, which demands a better memory allocator.

% \para{Garbage-collecting GPU tensors destabilizes the training speed.}

\begin{figure}[t]
    \centering
    \includegraphics[width=0.9\linewidth]{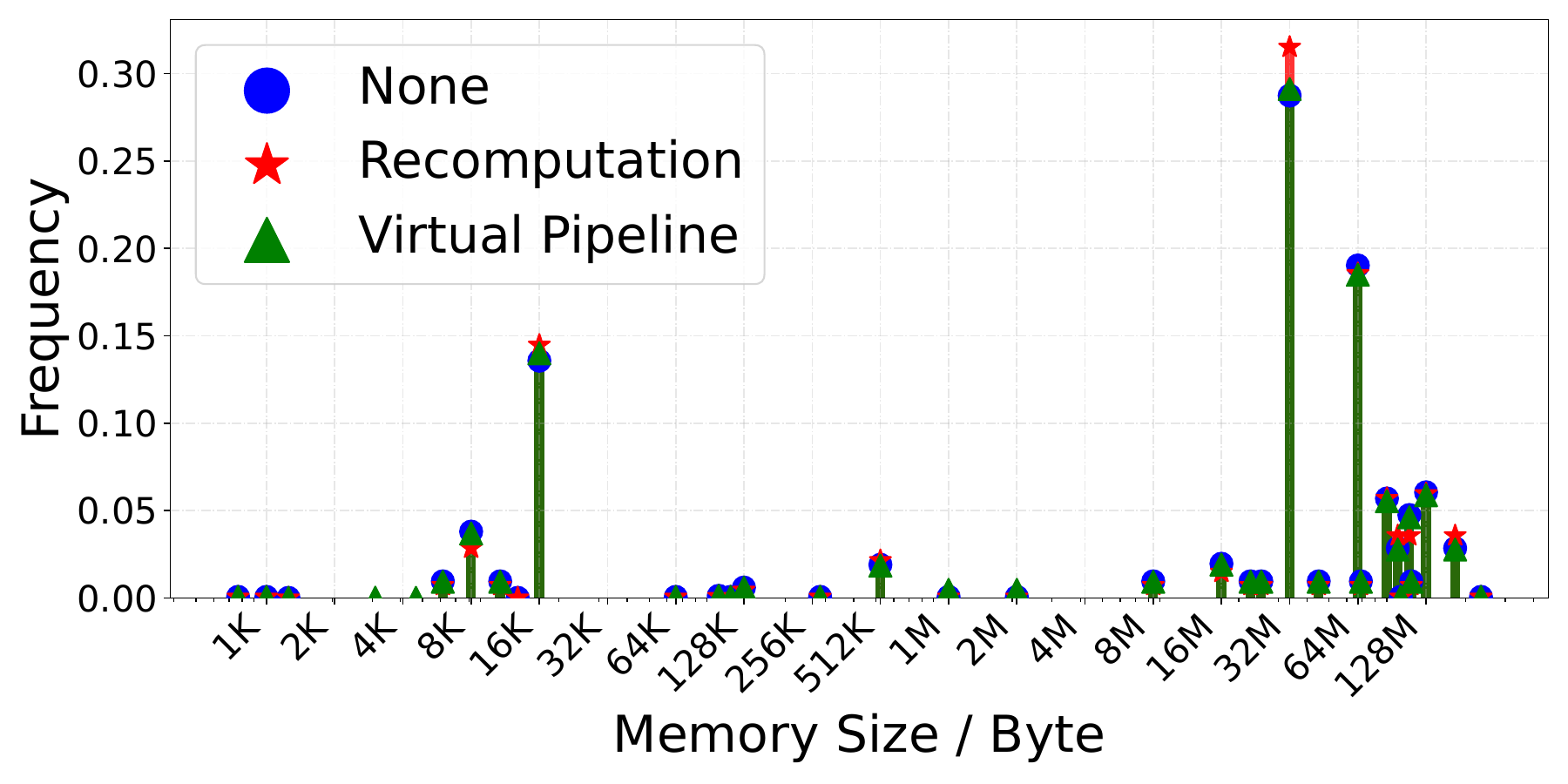}
    % \vspace{-10pt}
    \caption{Allocation size distribution during training. As shown in the figure, there are only around 32 distinct tensor sizes among different training configurations.}
    \label{fig:mem_alloc_size_distrib}
    \vspace{15pt}
\end{figure}

\begin{figure}[t]
    \centering
    \includegraphics[width=0.9\linewidth]{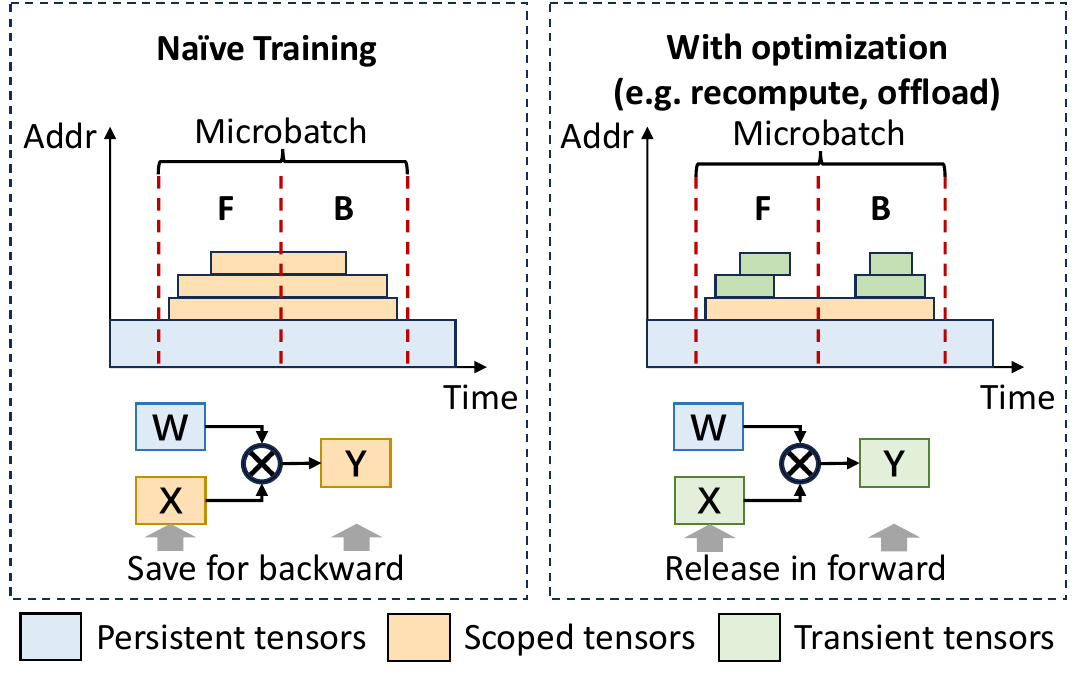}
    % \vspace{-15pt}
    \caption{Allocation classification based on temporal characteristic. The temporal characteristic of activation tensors are influenced by training optimization techniques.}
    \label{fig:mem_lifespan}
    % \vspace{-15pt}
\end{figure}
\subsection{Memory Behavior Insights in LLM Training}
\label{sec:regularity}

Low memory efficiency stems from complex allocation and deallocation requests, making it difficult for online allocators to minimize fragmentation. 
While defragmentation techniques such as block merging~\cite{pytorch} and virtual memory stitching~\cite{gmlake} can help, it is either suboptimal or introduce performance overhead (see \S\ref{sec:eval_overhead}). 
Fortunately, large model training presents an opportunity to address this challenge. 
We observe that it generates a largely predictable and periodic pattern of allocation requests in both spatial and temporal dimensions, which we term {\em allocation regularity}. 
Although optimization techniques such as virtual pipeline and recomputation add complexity, the overall allocation behavior remains regular. 
This regularity can be proactively exploited by allocators to create low-fragmentation plans in advance. 
Notably, we identify regularity across both spatial and temporal dimensions, as detailed below.

\noindent
\textbf{Spatial Regularity.}
Modern large models are comprised of a stack of Transformer layers or identical sub-networks~\cite{openai-gpt2}.
Consequently, the size of activation tensors generated during a training iteration exhibits significant repetition,
    which we call {\em spatial regularity}.
As shown in the Figure \ref{fig:mem_alloc_size_distrib}, 
    among over 50,000 tensor allocations with $>$512-byte size in a single training iteration of Llama2-7B, 
    there are only 32 distinct tensor sizes. 
Notably, with optimizations like recomputation and virtual pipeline,
    the regularity still persists---around 32 different sizes for $>$512-byte tensor allocations.

\noindent
\textbf{Temporal Regularity.}
We observe that tensor lifespans during language model training exhibit regular patterns, which can be categorized into three types as shown in Figure~\ref{fig:mem_lifespan}. 
\emph{Persistent tensors}, such as model weights, gradients, and optimizer states, are allocated at the beginning of training and remain in GPU memory throughout the training process. 
\emph{Scoped tensors} are allocated in one computation phase (the forward pass or backward pass of one microbatch) and released in another.
This type of tensor is mainly activation tensors of
forward computation and is used in backward computation.
As shown in Figure~\ref{fig:mem_lifespan}, scoped tensors are allocated sequentially in the forward computation and released in reverse order during the backward pass. 
\emph{Transient tensors}, such as intermediate input to unary operators (e.g., \texttt{ReLU}, \texttt{swiglu}), and activation tensors when training with optimization techniques like recomputation and offload, have very short lifespans and are released immediately after use, as they are not needed for backward computation. 
These temporal regularities can be effectively exploited in memory pre-planning to reduce inefficiencies caused by online decisions of allocation.

\section{\system Design Overview}
\label{sec:design}
\label{sec:workflow}

\begin{figure}[t]
    \centering
    \includegraphics[width=0.9\linewidth]{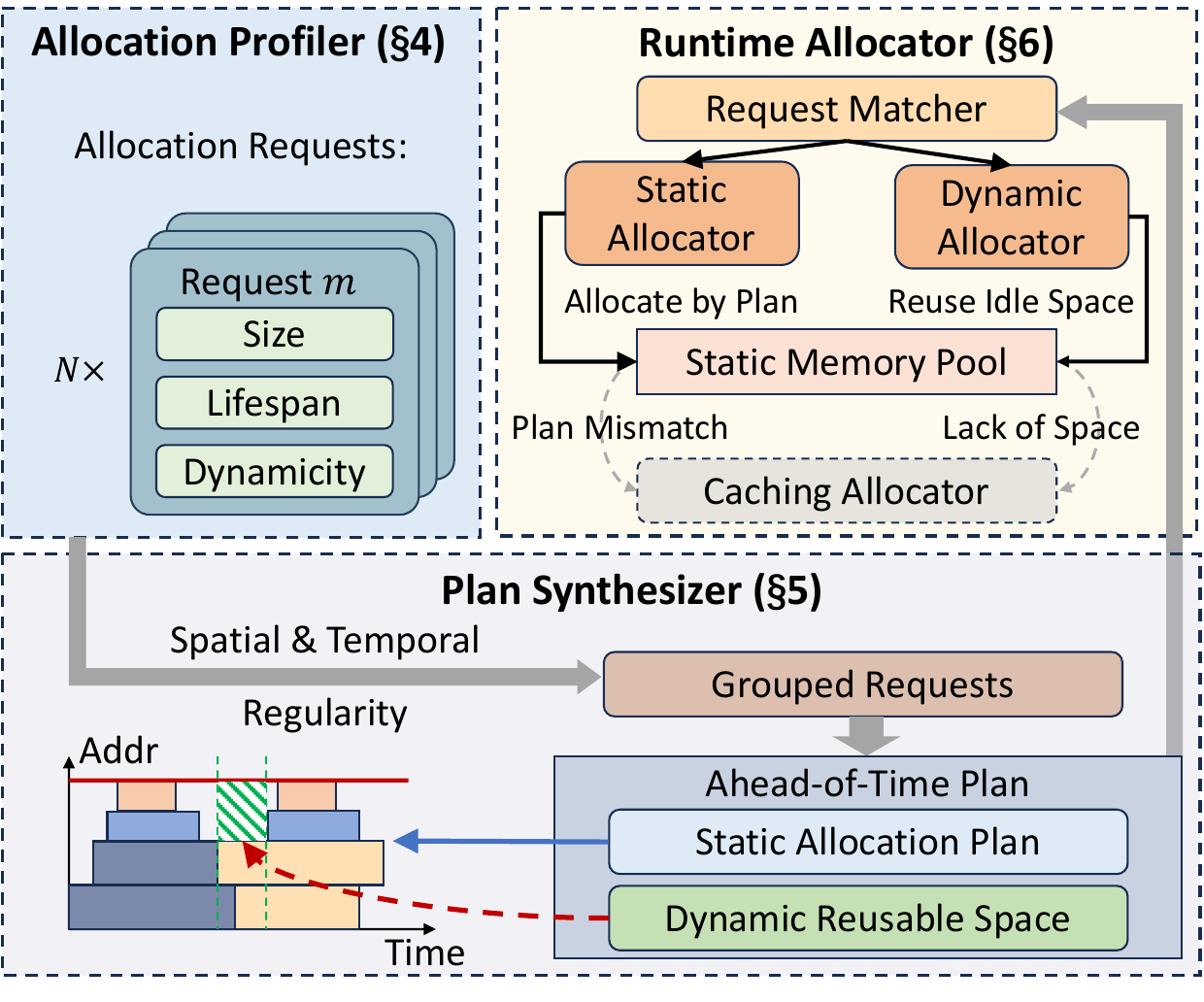}
    \vspace{-10pt}
    \caption{Workflow of \system.}
    \label{fig:workflow}
    \vspace{-15pt}
\end{figure}

\system comprises three components (Figure~\ref{fig:workflow}): Allocation Profiler (\S \ref{sec:profiler}), Plan Synthesizer (\S \ref{sec:plan_synthesizer}), and Runtime Allocator (\S \ref{sec:runtime_allocator}).
To generate an ahead-of-time GPU memory allocation plan, the initial step is to use the Allocation Profiler to capture the temporal (lifespan), spatial (size), and dynamicity information of all memory (allocation or free) requests within a training iteration. 
The request information is then fed to the Plan Synthesizer to generate an allocation plan. 
To this end, the Plan Synthesizer first groups the requests to reduce planning complexity based on their spatio-temporal regularities.
For static requests with fixed allocation size and lifespan, a {\em Static Allocation Plan} that minimizes memory fragmentation is generated leveraging the grouping results.
To handle dynamic requests with unpredictable allocation pattern, the Plan Synthesizer finds idle spaces (termed {\em Dynamic Reusable Space}) within the {\em Static Allocation Plan} that can be reused by dynamic requests later at runtime to further reduce fragmentation.
During training, the Runtime Allocator is used to perform the actual memory allocation, which consists of a Static Allocator and a Dynamic Allocator. 
The Static Allocator handles static requests based on the {\em Static Allocation Plan}, while the Dynamic Allocator attempts to allocate dynamic requests within the {\em Dynamic Reusable Space} if possible.
For dynamic requests that cannot be accommodated by the {\em Dynamic Reusable Space}, and any unexpected requests, the Runtime Allocator falls back to a caching allocator.

% \vspace{-10pt}
\section{Allocation Profiler}
\label{sec:profiler}
As described in \S\ref{sec:design}, 
    the Allocation Profiler traces each torch-level memory allocation and free request to capture its spatial, temporal, and dynamicity information.
Notably, apart from the basic information like request timestamp, address, size, and dynamicity, 
    the Allocation Profiler also records training-level information including the current computation phase (forward or backward), micro-batch ID, and the module name that issues the request to facilitate the identification of spatio-temporal regularities.

Formally, we organize an allocation request and its associated free request into a memory request event $m$, which is defined as $m := (s, t_s, t_e, p_s, p_e, dyn)$. 
Here, $s$ represents the request size; $t_s$ and $t_e$ are the allocation and free timestamps of the memory chunk, respectively; 
    $p_s$ and $p_e$ identify the computation phases of allocation and free, respectively; 
    and $dyn$ is a boolean flag indicating if the request originates from a dynamic layer (e.g., a MoE expert layer). 
For requests from dynamic layers (where $m.dyn=True$), two additional elements $l_s$ and $l_e$ are recorded, which are the originating module name for the allocation and free, respectively. 
This additional information allows us to group dynamic requests based on their temporal regularity, further details are provided in \S\ref{sec:dynamic-planning}.
Upon completion, the profiler outputs a list $\mathcal{M}$ of these characterized allocation requests, which is the primary input of the Plan Synthesizer.

\begin{figure*}[t]
    \centering
    \includegraphics[width=0.97\linewidth]{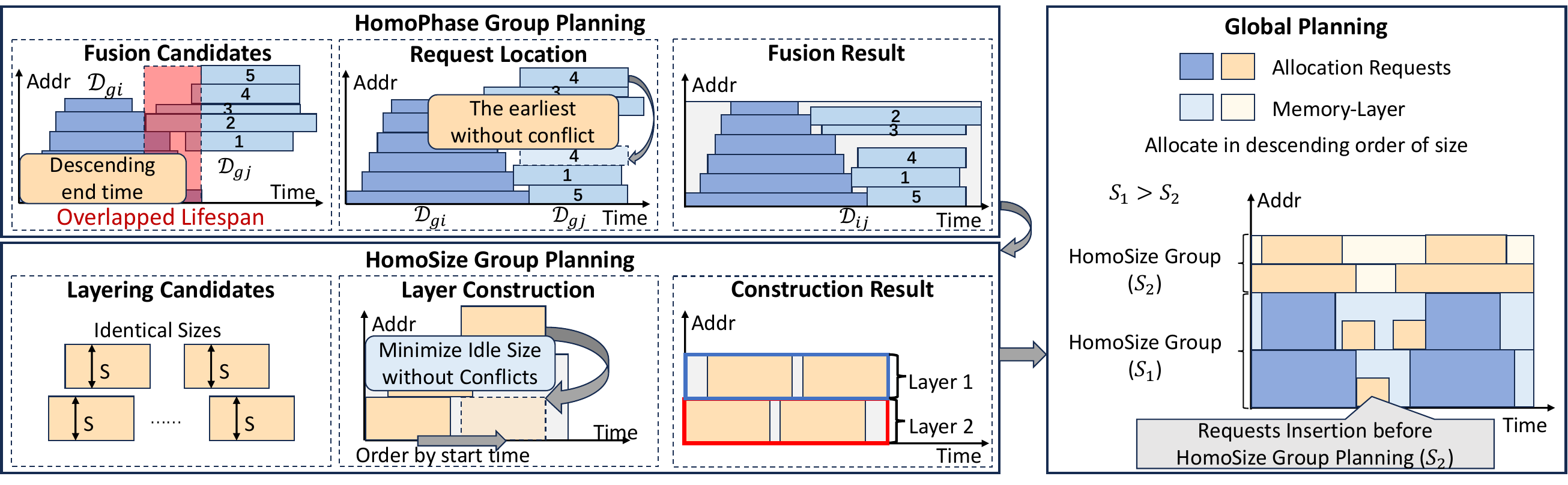}
    \vspace{-10pt}
    \caption{Static Allocation Planning of allocation requests. Allocation requests are first grouped based on temporal characteristics into {\em HomoPhase Groups} for intra-group planning (upper left), and then further grouped based on spatial characteristics into {\em HomoSize Groups} (bottom left). During global planning, {\em HomoSize Groups} are inserted and placed in descending order of allocation size (right).}
    \label{fig:group_fusion}
\end{figure*}

\section{Plan Synthesizer}
\label{sec:plan_synthesizer}
The goal of the plan synthesizer is to produce a low fragmentation memory allocation plan that maximizes memory efficiency $E$ as defined in \S\ref{sec:low-mem-eff}.
Since allocated memory $M_a$ is fixed for a specific training configuration, the goal is then to minimize reserved memory $M_r$.
% Formally, the input of the synthesizer is a subset of the profiled allocation requests $\mathcal{M}$,
%     denoted as $\mathcal{M}_s$, 
%     which excludes allocation requests marked as dynamic by the profiler, 
%     i.e. $\mathcal{M}_s := \{m | m \in \mathcal{M}, m.d = False\}$.
% We handle dynamic allocations at run time as to be described in \S\ref{sec:runtime_allocator}.
To this end, the input of the synthesizer $\mathcal{M}$ is first partitioned into two subsets $\mathcal{M}_s := \{m | m \in \mathcal{M}, m.dyn = False\}$ and $\mathcal{M}_d := \{m | m \in \mathcal{M}, m.dyn = True\}$ according to their dynamicity.
For $\mathcal{M}_s$ containing static request events, we perform static allocation planning to generate the {\em Static Allocation Plan}. 
Next, for dynamic request events $\mathcal{M}_d$, we find idle space in the plan (called {\em Dynamic Reusable Space}) that can be used to handle dynamic requests at runtime.
% Based on the lifespan properties, we group the dynamic requests and identify reusable address ranges, called {\em Reuseable Idle Address} within their corresponding time windows.
% These identified regions serve as guidance for subsequent runtime allocations.
% \vspace{-10pt}
\subsection{Static Allocation Planning}
The {\em Static Allocation Plan}, denoted as $\mathcal{D}_s$, consists of a list of allocation decisions.
Each allocation decision $d \in \mathcal{D}_s$ incorporates the six attributes of $m$ and is augmented with an additional attribute, $a$, which denotes the start address of the allocated memory chunk, i.e., $d:=m + (a) = (s, t_s, t_e, p_s, p_e, dyn, a)$.
% \zixiao{Move forward to Sec 2} Following prior work~\cite{gmlake},
%     we calculate the memory efficiency $U$ during a period of time to be $U = \frac{A}{R}$,
%     where $A$ and $R$ are the maximum total size of allocated memory and
%     the maximum total size of reserved memory during the time, respectively.
The allocation planning is then under the constraint that for any two allocation decisions $d_i$ and $d_j$, 
    they cannot simultaneously have conflicting lifespans and conflicting address ranges.
Otherwise, they will have intersecting memory and result in memory stomping.

Since finding the optimal allocation plan is NP-hard and involves a large input scale as described in \S\ref{sec:intro}, brute-force methods or any pruning techniques that do not fundamentally reduce the complexity of the original search space~\cite{telamalloc, minimalloc} are infeasible.
Inspired by the spatio-temporal regularity we uncover in \S\ref{sec:regularity},
    our idea is to decouple the searching in the space (i.e., memory address and size) and time (allocation time and free time) dimensions.
In the global planning process, requests that exhibit temporal or spatial regularities are handled through local planning, where efficient layouts are derived by exploiting these regularities. 
The resulting local plans then become integral components of the global allocation plan. 
To support this workflow, we introduce two abstractions: {\em HomoPhase Group} and {\em HomoSize Group}.
% , which classify allocation requests by temporal and spatial characteristics, respectively, and guide the local planning.
The {\em HomoPhase Groups} gather allocation requests with similar lifespan, and the {\em HomoSize Group} gather requests with same size.
We will further introduce them in the following.
% Therefore, we propose two abstractions of {\em HomoPhase Group} and {\em HomoSize Group} to gather allocation requests with the same characteristics in temporal and spatial, respectively.
% These two abstractions serve as the objects of the local planning and also constitute the building parts of the global planning.

% At a high level, we first group memory allocation requests with the same temporal or spatial characteristics, devise a local plan for them to reduce the problem space,
%     and then perform a global planning to reach the final plan.
% During the local and global planning, we greedily combine the optimal local memory layout for each request size to approximate the global optimal solution.
% We devise special abstractions for the temporal and spatial groups in the above algorithmic workflow,
%     namely {\em HomoPhase Group} and {\em HomoSize Group}, respectively.

\para{Global Planning.}
% In the local planning, memory requests are grouped based on their temporal and spatial regularities into {\em HomoPhase Groups} and {\em HomoSize Groups}, and generate a local memory allocation plan for each group. 
% These local plans can then be regarded as building blocks of the global memory allocation plan, as shown in Figure~\ref{fig:group_fusion}.
At a high level, the components of the global planning are derived through local plans. 
Each local plan groups allocation requests that exhibit regularity along one dimension (spatial or temporal), allowing us to exploit such regularities to design allocation algorithms that improve memory efficiency.

Specifically, the first step in generating the global allocation plan is to partition all memory requests within one training iteration into different {\em HomoPhase Groups} based on their temporal characteristics. 
Adjacent {\em HomoPhase Groups} are then merged to produce local plans, denoted as $\mathcal{D}_g$ (see in {\em HomoPhase Group} Planning, Figure~\ref{fig:group_fusion} upper left).
These local plans are subsequently treated as unified memory requests in the next stage of spatial grouping, where they are classified into different {\em HomoSize Groups} according to their allocation sizes. 
Each {\em HomoSize Group} will construct its own local plan.
We execute the construction in descending order of the request size for the {\em HomoSize Group}, since smaller memory requests may fit into the unused intervals of larger requests, thus improving overall memory efficiency.
% Therefore, before planning for {\em HomoSize Group} of size $S_i$ (at which point all larger groups have already been processed), we first attempt to insert the requests of size $S_i$ into the free intervals of existing local plans for {\em HomoSize Group} of size larger than $S_i$.
Before planning for a {\em HomoSize Group} of size $S_i$, all larger groups have already been processed. 
At this stage, we first try to place the requests of size $S_i$ into the free intervals of the existing local plans of larger groups.
The remaining requests of size $S_i$ that cannot be placed in the larger group plans will then construct a new local plan (see in {\em HomoSize Group} Planning, Figure~\ref{fig:group_fusion} bottom left).
Finally, after all local plans for {\em HomoSize Group} have been constructed, each memory request is assigned a specific address within the global allocation plan (Figure~\ref{fig:group_fusion} right).

Next, we introduce how local plans for {\em HomoPhase Group} and {\em HomoSize Group} are generated.

\begin{figure*}[t]
    \centering
    \includegraphics[width=0.95\linewidth]{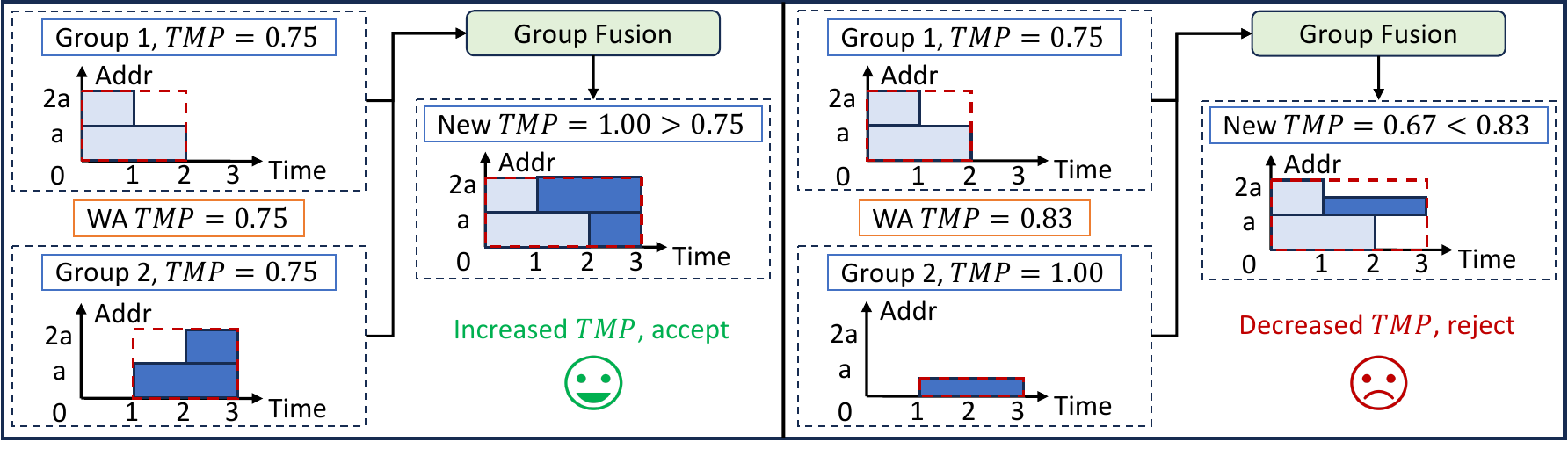}
    \caption{Examples of fusion between different {\em HomoPhase Groups}. In the left example, the fusion leads to an increase in $TMP$ (`WA' means weighted average), so the fusion scheme is adopted; in the right example, the fusion results in a decrease in $TMP$, so it is rejected.}
    \label{fig:fusion_example}
\end{figure*}

% \noindent
\para{HomoPhase Group Planning.}
A {\em HomoPhase Group} $\mathcal{M}_g$ contains allocation requests that start and end in the same computation phases, which is:
$
\mathcal{M}_g := \{m \in \mathcal{M}_s \mid m.p_s = P_s,\ m.p_e = P_e\}.
$
Here, $P_s$ and $P_e$ denote a pair of computation phases (e.g., forward/backward passes), meaning all requests in $\mathcal{M}_g$ share similar lifespans.
$\mathcal{D}_g$ is the allocation plan for {\em HomoPhase Group} $\mathcal{M}_g$, where each allocation request are placed with a relative address.

Since their lifespans overlap, packing them contiguously into a single memory block achieves local optimal. 
However, their lifespans are only partially aligned, some memory may remain unused during parts of the timeline. These gaps are called {\em spatio-temporal bubbles}, causing memory fragmentation.

To reduce such bubbles, we fuse adjacent groups when the end phase of one matches the start phase of another. The merged group can better reuse memory across phase boundaries. We evaluate memory efficiency using the {\em time-memory product} ($TMP$)~\cite{best-fit-allocation}:
\begin{align}
% \small
\begin{split}
    TMP &= \frac{\sum\limits_{d \in \mathcal{D}_g} d.s \cdot (d.t_e - d.t_s)}{\mathcal{D}_g.s \cdot (\mathcal{D}_g.t_e - \mathcal{D}_g.t_s)}, \\
    \mathcal{D}_g.s &= \max_{d \in \mathcal{D}_g} (d.a + d.s), \\
    \mathcal{D}_g.t_s &= \min_{d \in \mathcal{D}_g} d.t_s, \quad \mathcal{D}_g.t_e = \max_{d \in \mathcal{D}_g} d.t_e.
\end{split}
\end{align}
The numerator measures actual memory-time usage; the denominator reflects the reserved memory over time. A higher $TMP$ indicates fewer bubbles.

As shown in Figure~\ref{fig:group_fusion} upper left, we fuse two local plans $\mathcal{D}_{gi}$ and $\mathcal{D}_{gj}$ by inserting the smaller one into the larger. Assume $\mathcal{D}_{gi}.s > \mathcal{D}_{gj}.s$; we sort $\mathcal{D}_{gi}$ by end time in descending order and try placing each $d_j \in \mathcal{D}_{gj}$ at the lowest available address $addr$, starting from:
$
addr = \min_{d_i \in \mathcal{D}_{gi}} d_i.a.
$
At each step:

1. Choose the earliest-starting $d_j$ that fits without conflict and place it at $addr$. Update $addr \leftarrow addr + d_j.s$.

2. If no fit is found, move $addr$ to the next $d_i.a$ in $\mathcal{D}_{gi}$:
\[
addr \gets \min_{d_i \in \mathcal{D}_{gi},\ d_i.a > addr} d_i.a.
\]

The fusion is accepted only if the new $TMP$ increases over the weighted average of the originals, meaning fewer bubbles, as shown in Figure~\ref{fig:fusion_example}.

Each (possibly fused) group then forms a local plan $\mathcal{D}_g$, where requests are given relative addresses. We treat this plan as a single large request $m_g$ for global planning:
\[
m_g.s = \mathcal{D}_g.s, \quad m_g.t_s = \mathcal{D}_g.t_s, \quad m_g.t_e = \mathcal{D}_g.t_e.
\]

\para{HomoSize Group Planning.}
\begin{algorithm}[t]
\small
    \caption{Memory-Layer Construction for {\em HomoSize Group}.}
    \label{algo:layer_cons}
    \SetAlgoLined
    \SetKwData{Left}{left}\SetKwData{This}{this}\SetKwData{Up}{up}
    \SetKwFunction{Union}{Union}\SetKwFunction{FindCompress}{FindCompress}
    \SetKwInOut{Input}{Input}\SetKwInOut{Output}{Output}
    \Input{Request Set $\mathcal{M}_s=\{m|m.s=S, m \in \mathcal{M}\}$}
    \Output{Memory-Layer List $\mathcal{L}=(\mathcal{M}_{l1}, \mathcal{M}_{l2}, ...)$}
    \BlankLine
    $\mathcal{L}\leftarrow[\:]$\;
    $\mathcal{M}_s$.sort(key=$m.t_s$)\;\label{line:sort}
    \For{$m\in\mathcal{M}_s$}{
    $\mathcal{M}_{l}\leftarrow\max\limits_{L.end}\{L \in \mathcal{L} : L.end < m.t_s\}$\; \label{line:find_layer}
    \If{$\mathcal{M}_{l}$ is $None$}{\label{line:if_start}
    $\mathcal{M}_{l}\leftarrow$ new Memory-Layer(size=$m.s$)\;
    $\mathcal{L}$.append($\mathcal{M}_{l}$)\;
    }\label{line:if_end}
    $\mathcal{M}_{l}$.append($m$)\;
    $\mathcal{M}_{l}.end\leftarrow m.t_e$\;
    }
\end{algorithm}

Allocation requests exhibit strong repetitiveness, with many requests having identical allocation sizes and differing only in their lifespan.
This observation continues to hold true after {\em HomoPhase Group} planning, primarily because each microbatch exhibits identical behavior during training. 
% Consequently, the requests encompassed within the computation phases of different microbatches are consistent in both quantity and size. 
Therefore, the {\em HomoPhase Group} formed through temporal grouping and fusion also possesses the characteristic that multiple such groups are identical in size, differing only in their lifespan.

Based on this observation, we propose the abstraction termed {\em HomoSize Group},
    which aggregates allocation requests of the same size property.
For requests of a specific size $S$, there are only differences in their lifespan.
Therefore, any subset of these requests with non-overlapping lifetimes can reuse the same space in GPU memory. 
In the time-space coordinate system, this shared space can be regarded as a layer within the memory space, referred to as a memory-layer.
To obtain a local optimal allocation plan for memory requests within a {\em HomoSize Group}, we need to minimize the number of memory-layers required to allocate all requests.

Algorithm~\ref{algo:layer_cons} describes the procedure of constructing memory-layers for {\em HomoSize Groups} of a specific size $S$.
To begin with, the allocation requests of size $S$ are included in a {\em HomoSize Groups} $\mathcal{M}_s$, 
    and are sorted by their allocation time (Line~\ref{line:sort}).
Next, for each allocation request $m \in \mathcal{M}_s$, 
    we try to find a memory-layer whose last allocation request's free time is closest to but smaller than $m$'s allocation time,
    so as to minimize the idle time of the memory-layer while avoiding conflicting lifespans (Line~\ref{line:find_layer}).
If we can find such a memory-layer,
    $m$ is appended to the layer's tail.
Otherwise, a new memory-layer is constructed and is populated by $m$ (Line~\ref{line:if_start} - \ref{line:if_end}).
In this way, we minimize both the intra-layer gaps and the total number of memory-layers.

\subsection{Locating Dynamic Reusable Space}
\label{sec:dynamic-planning}
Dynamic allocation requests ($\mathcal{M}_d$) are characterized by sizes determined only at runtime, necessitating online allocation. 
A key insight from our profiler analysis is that the peak memory usages of static and dynamic allocations typically do not occur simultaneously.
% Therefore, managing static and dynamic allocations in separate memory address regions, where each region is provisioned for its respective peak, results in memory fragmentation. 
% For instance, when dynamic memory usage peaks, the static memory region often contains substantial idle capacity.
% This observation of underutilized memory, a direct consequence of misaligned peak demands under segregated management, forms the primary motivation for our proposed dynamic planning strategy.
% If they are managed in separate memory address regions, each region must be provisioned for its own peak demand. 
% As a result, substantial portions of memory remain idle whenever the other type of allocation dominates, leading to reduced overall memory efficiency. 
% This inefficiency, arising from the temporal misalignment of static and dynamic peaks, motivates our dynamic planning strategy, which is reusing the idle space of static allocations to serve dynamic memory requests.
When static and dynamic allocations are managed in separate memory regions, each region must be provisioned for its individual peak usage. 
However, since the peaks of static and dynamic demands occur at different times, the reserved capacity of one region (e.g., static) often remains idle while the other (e.g., dynamic) reaches its peak. 
This temporal mismatch results in significant underutilization of memory resources and lowers overall efficiency. 
To overcome this limitation, our dynamic planning strategy reuses the idle space of static allocations to accommodate dynamic requests.

To improve memory efficiency and reduce peak memory consumption during training, dynamic requests should reuse idle spaces within the {\em Static Allocation Plan} as much as possible. 
However, allocating dynamic requests directly within the available spaces of the {\em Static Allocation Plan} at runtime may lead to memory stomping. 
This occurs because the current dynamic allocation request might overlap in address space with subsequent static allocations that are already planned.
We observe that, although the sizes of dynamic allocations are unpredictable, their lifespan are relatively fixed. 
Leveraging this temporal regularity, we can identify reusable regions within the {\em Static Allocation Plan} before training, providing guidance for online allocation at runtime.

Leveraging the predictable lifetimes of dynamic memory allocations, our approach proactively identifies reusable memory regions within the {\em Static Allocation Plan} before runtime.
In contrast to the computation-phase granularity used for static allocations, here we operate at the model layer level to achieve fine-grained temporal precision for these dynamic requests. 
This refined granularity enables a more precise interrogation of {\em Dynamic Reusable Space}. 
% Such accurately identified regions within the typically shorter layer-level intervals tend to be more effectively utilized, maximizing opportunities for dynamic memory reuse and thereby reducing the peak GPU memory footprint.
Since these regions lie within shorter layer-level intervals, they can be utilized more effectively, creating more opportunities for dynamic memory reuse and thereby lowering the peak GPU memory footprint.
We characterize each dynamic request by its malloc model layer, $l_s$, and its free model layer, $l_e$ (profiling methodology in \S\ref{sec:profiler}). 
This ($l_s$, $l_e$) pair establishes a bounding temporal interval, from $l_s$'s start to $l_e$'s end, which contains the lifespan of the dynamic allocation.
To systematically manage these lifetimes, we classify all dynamic allocation requests into distinct groups, called {\em HomoLayer Group}, where each group $\mathcal{G}$ comprises requests sharing identical ($l_s$, $l_e$) pairs:
\begin{equation}
% \small
    \mathcal{G}(a, b):=\{m|m.l_s=a, m.l_e=b\}
\end{equation}
where $a$ and $b$ represent for the dynamic layers in the model.
For every such group of dynamic requests $\mathcal{G}(a, b)$, and its corresponding temporal range $\mathcal{T}(a, b)=[a.start,b.end]$, we then interrogate the pre-established {\em Static Allocation Plan} $\mathcal{D}_s$. 
The objective of this interrogation is to identify all contiguous memory segments that remain idle throughout the entirety of this specific temporal range. 
In the {\em Static Allocation Plan} $\mathcal{D}_s$, each decision $d$ contains a static allocation request $m$ and its allocate address $a$, indicating the spatial and temporal occupation space for $d$ is $R_s(d)=[d.a, d.a+d.s]$ and $R_t(d)=[d.t_s, d.t_e]$ respectively. 
The occupied address ranges for $\mathcal{T}(a, b)$ can be represented as:
\begin{equation}
% \small
    \mathcal{A}_o (a, b)=\underset{d\in \mathcal{D}_s,\ R_t(d)\cap \mathcal{T}(a, b) \neq \emptyset}{\cup} R_s(d)
\end{equation}
The {\em Dynamic Reusable Space} $\mathcal{A}_i$ ranges during $\mathcal{T}(a, b)$ are the complement of all addresses $\mathcal{A}$ occupied in the allocation plan, as shown in Eq.~\ref{eq_a} and Eq.~\ref{eq_ai}.
\begin{align}
% \small
    &\mathcal{A}=[\min_{d \in \mathcal{D}_s} (d.a), \max_{d \in \mathcal{D}_s} (d.a+d.s)] \label{eq_a} \\
    &\mathcal{A}_i (a, b)=\mathcal{A}\setminus \mathcal{A}_o (a, b) \label{eq_ai}
\end{align}
The identified {\em Dynamic Reusable Space} $\mathcal{A}_i$ are subsequently designated as candidate reusable regions. 
At runtime, when a dynamic allocation request belonging to a particular ($l_s$,$l_e$) group arises, the allocator can preferentially utilize these pre-vetted regions, thereby ensuring that dynamic allocations are placed in memory spaces that will not conflict with future, planned static allocations.

% \vspace{-10pt}
\section{Runtime Allocation}
\label{sec:runtime_allocator}

The runtime allocator manages the GPU memory and serves allocation requests based on the allocation plan generated by the plan synthesizer.
It consists of two main components, a static allocator that handles allocation requests {\em without} runtime dynamics (i.e., $m.dyn == False$),
    and a dynamic allocator that handles allocations {\em with} runtime dynamics (i.e., $m.dyn == True$).
During runtime, when an allocation request is received by the Runtime Allocator, the Request Matcher routes the request to an appropriate allocator based on the dynamic characteristics of the current model layer (detail shows in \S\ref{sec:implementation}).
Furthermore, to address scenarios such as potential mismatch between actual runtime allocation requests and the {\em Static Allocation Plan}, or instances of inadequate {\em Dynamic Reusable Space} for dynamic requests, \system's runtime allocation further incorporates a caching allocator. 
This component is designed to manage these exceptional cases, thereby guaranteeing the overall robustness of the system.
% \vspace{-10pt}
\subsection{Static Allocator}
The static allocator, guided by the {\em Static Allocation Plan}, reserves a static memory pool prior to training, where the size of the memory pool is fixed, defined by the result of {\em Static Allocation Plan}.
At runtime, it efficiently serves static requests by providing pre-planned memory addresses sequentially. 
This eliminates the need for online allocation searches found in systems like PyTorch. 
% \vspace{-10pt}
\subsection{Dynamic Allocator}
Certain models, such as the Mixture-of-Experts (MoE), exhibit non-deterministic memory patterns, making it impossible to pre-plan memory addresses for all tensors. To handle these cases, we employ a dynamic allocator that assigns memory at runtime.

The primary strategy of the dynamic allocator is to prioritize reusing memory from the static memory pool, which was pre-allocated for predictable requests. To prevent conflicts, \system meticulously tracks all currently available address intervals ($\mathcal{A}_a$) in this pool. When any memory block is allocated or freed, $\mathcal{A}_a$ is updated accordingly.

When a dynamic request $m$ arrives, the allocation process begins by first identifying the {\em Dynamic Reusable Space} $\mathcal{A}_i$, which is the available space in static memory pool for the {\em HomoLayer Group} contains $m$. 
Since prior allocations may have already occupied parts of this space, we must identify the portions that are still free.
To find the actual memory available for allocation, \system computes the intersection of this potential space $\mathcal{A}_i$ with the currently free intervals $\mathcal{A}_a$. This calculation yields a set of candidate intervals, $\mathcal{A}_c(m)$:
\begin{equation}
    \mathcal{A}_c(m) = \mathcal{A}_a \cap \mathcal{A}_i
    \label{eq:candidate_intervals}
\end{equation}
From these candidate intervals, we apply the best-fit policy to select the most suitable one for the request. Once an interval is chosen and the memory is assigned, the system updates the list of available intervals $\mathcal{A}_a$ to reflect the allocation.

If no candidate interval in the static pool can satisfy the request, the system falls back to the caching allocator as a secondary option.
This caching allocator follows PyTorch’s CUDA Caching Allocator, reusing previously allocated blocks through a block management mechanism.

\section{Complexity Analysis}
We analyze the computational complexity of our method. The discussion is organized into two main parts: 
(i) plan synthesis, which is performed once before execution, 
and (ii) runtime allocation, which is invoked during model training or inference.

\subsection{Plan Synthesis}
\noindent
\textbf{Static Requests.}
Plan synthesis for static requests consists of two stages. First, requests
are grouped into {\em HomoPhase Groups} and fused when possible, which requires sorting by request endpoints and costs $O(N \log m)$, where $N$ is the total number of requests and $m$ is the maximum group size. 
Second, the resulting groups are ordered and inserted by size. 
Sorting by size requires $O(N \log N)$, while the layer construction within each group
(Algorithm~\ref{algo:layer_cons}) is linear in group size. Thus, static plan synthesis has overall complexity
$O(N \log N)$.

\noindent
\textbf{Dynamic Requests.}
For dynamic requests, all $k$ {\em HomoLayer Groups}' temporal intervals are known in advance. 
By sorting time intervals and performing a batched sweep, the complexity is:
\[
O\big(N \log N + k \log(N+k) + \sum_{i=1}^k r_i\big),
\]
where $r_i$ is the number of static requests overlapping the $i$-th query.
Since typically $k \ll N$ and $\sum_i r_i \ll kN$, the dynamic part is asymptotically bounded by $O(N \log N)$.

\subsection{Runtime Allocation}
\noindent
\textbf{Static Requests.}
At runtime, static requests incur $O(1)$ lookup cost because their addresses are pre-computed in the plan.

\noindent
\textbf{Dynamic Requests.}
Dynamic requests are allocated by intersecting pre-computed reusable spaces
with the currently active blocks. If $n$ is the number of active blocks at
that moment, this step costs $O(n)$. In practice $n$ is orders of magnitude
smaller than $N$, ensuring that runtime allocation remains efficient.

Overall, plan synthesis is dominated by $O(N \log N)$ preprocessing cost,
while runtime allocation requires only $O(1)$ per static request and $O(n)$
per dynamic request. This guarantees that both initialization and runtime
operations are efficient in large-scale training.

% \vspace{-10pt}
\section{Implementation}
\label{sec:implementation}

\system is implemented for PyTorch using about 3100 lines of Python and 2300 lines of C/C++.
The plan synthesizer is implemented as a standalone tool,
    while the profiler and allocator are implemented as PyTorch's PluggableAllocator~\cite{pytorch-doc},
    which can be loaded before training to take over the \texttt{malloc} and \texttt{free} API calls.
This means that \system is compatible with any PyTorch version and GPU platform that supports the PluggableAllocator interface.
To capture temporal and spatial characteristics, \system employs monkey patching for lightweight instrumentation, requiring no more than five lines of code in the original training framework.

\noindent
\textbf{Allocation Profiler.}
The profiler is designed to log tensor allocation requests made by PyTorch-based model training frameworks. 
It interfaces directly with native GPU memory allocation APIs, such as {\tt cudaMalloc} and {\tt cudaFree} for NVIDIA GPUs. 
This approach ensures that memory is allocated precisely as required, thereby almost entirely obviating memory fragmentation under these conditions. 
Consequently, the profiler can trace GPU memory for training configurations that would lead to out-of-memory (OOM) errors with PyTorch's default allocator. 
If an OOM error occurs even when using these native GPU APIs for profiling, it indicates that the configuration's theoretical memory demand inherently surpasses the GPU's memory capacity, rendering it impossible to execute irrespective of fragmentation. 
% To capture spatio-temporal allocation details and information on computation phases and dynamic layers—crucial for analysis and plan generation—a few lines of code are added to the training framework, typically around four lines for frameworks like Megatron-LM, using monkey patching and PyTorch hook APIs~\cite{pytorch-doc}.

\noindent
\textbf{Runtime Allocator.}
At runtime, the allocator performs memory allocation according to the allocation plan. 
During training initialization, \system uses native GPU memory allocation APIs to preallocate a contiguous memory block equal in size to the static memory pool and also initializes a caching allocator as a fallback. 
The runtime allocator then assigns address ranges within the static memory pool without issuing additional GPU memory API calls, thereby avoiding extra runtime overhead.
To identify the current model layer during execution and route memory requests to the appropriate allocator, \system leverages PyTorch’s hook APIs to track the execution of model modules. 
When a memory request arrives at runtime, the Request Matcher in the Runtime Allocator uses the current module information to determine whether the request should be handled by the static allocator according to the allocation plan, or by the dynamic allocator for online allocation.
\begin{figure*}[t]
    \centering
    \begin{subfigure}[b]{0.3\textwidth}
        % \centering
        \includegraphics[width=\linewidth]{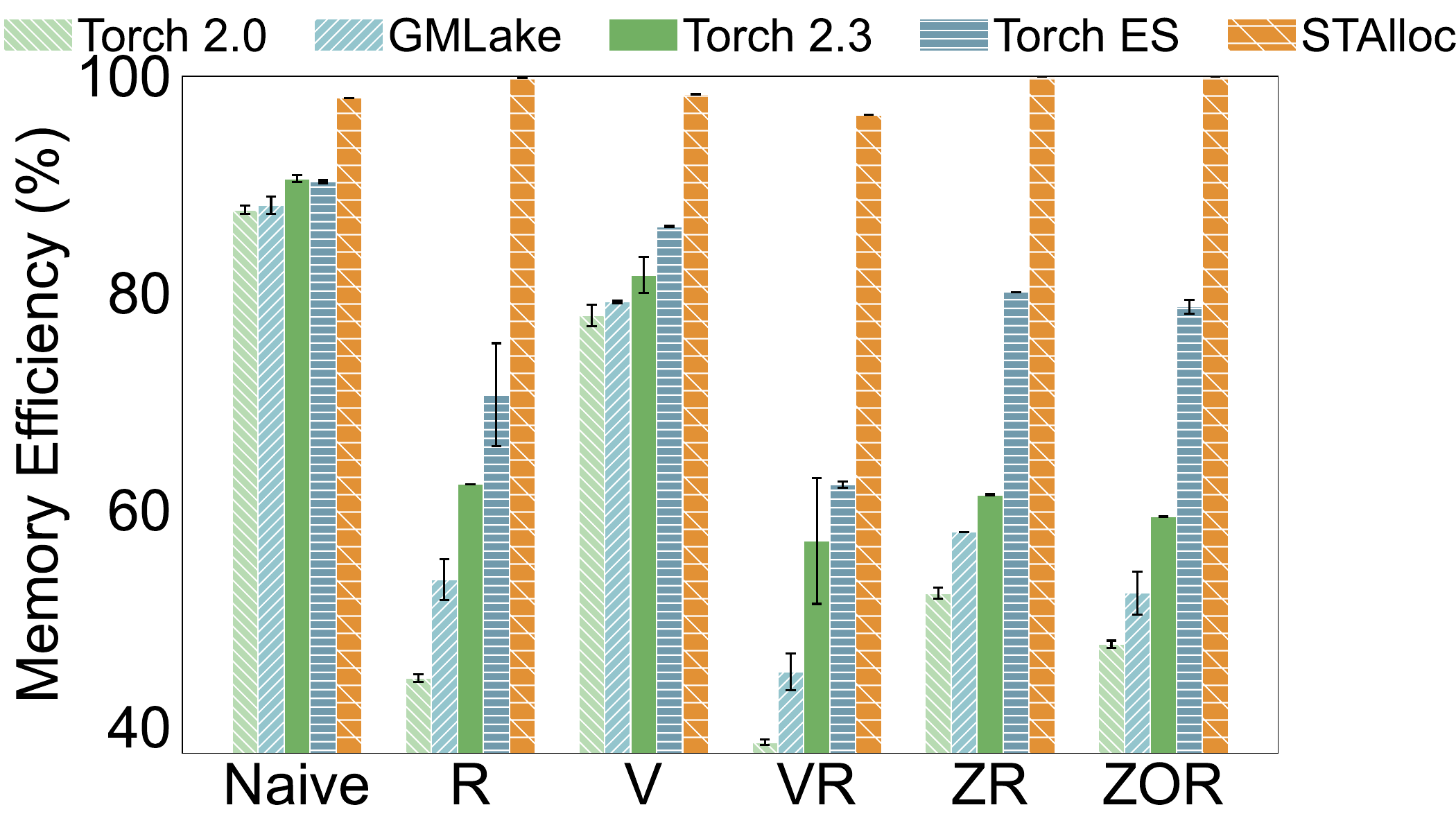}
        \vspace{-15pt}
        \caption{GPT-2.}
        \label{fig:a800_gpt2}
    \end{subfigure}
    \begin{subfigure}[b]{0.3\textwidth}
        % \centering
        \includegraphics[width=\linewidth]{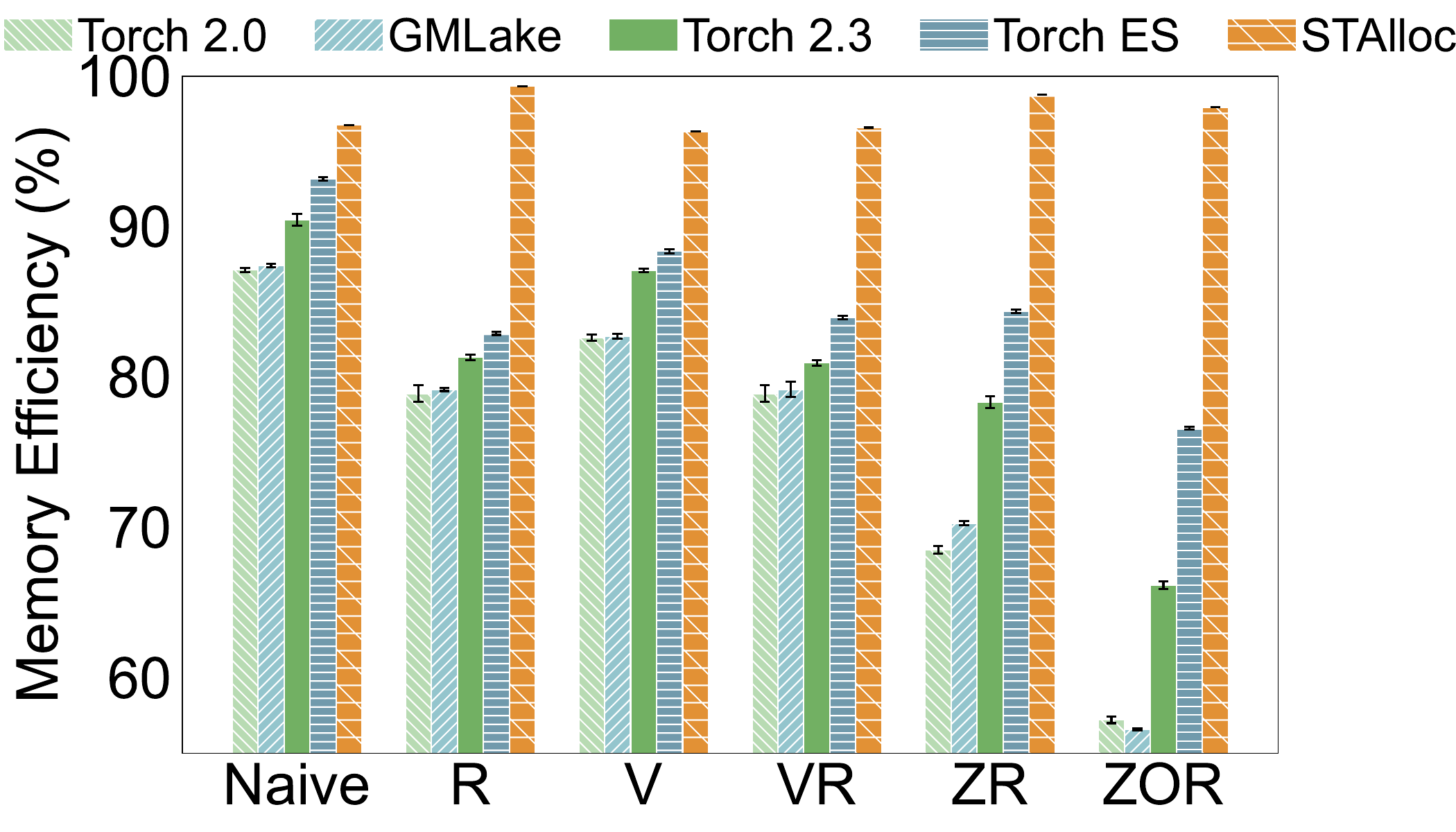}
        \vspace{-15pt}
        \caption{Llama2-7B.}
        \label{fig:a800_llama2}
    \end{subfigure}
    \begin{subfigure}[b]{0.3\textwidth}
        % \centering
        \includegraphics[width=\linewidth]{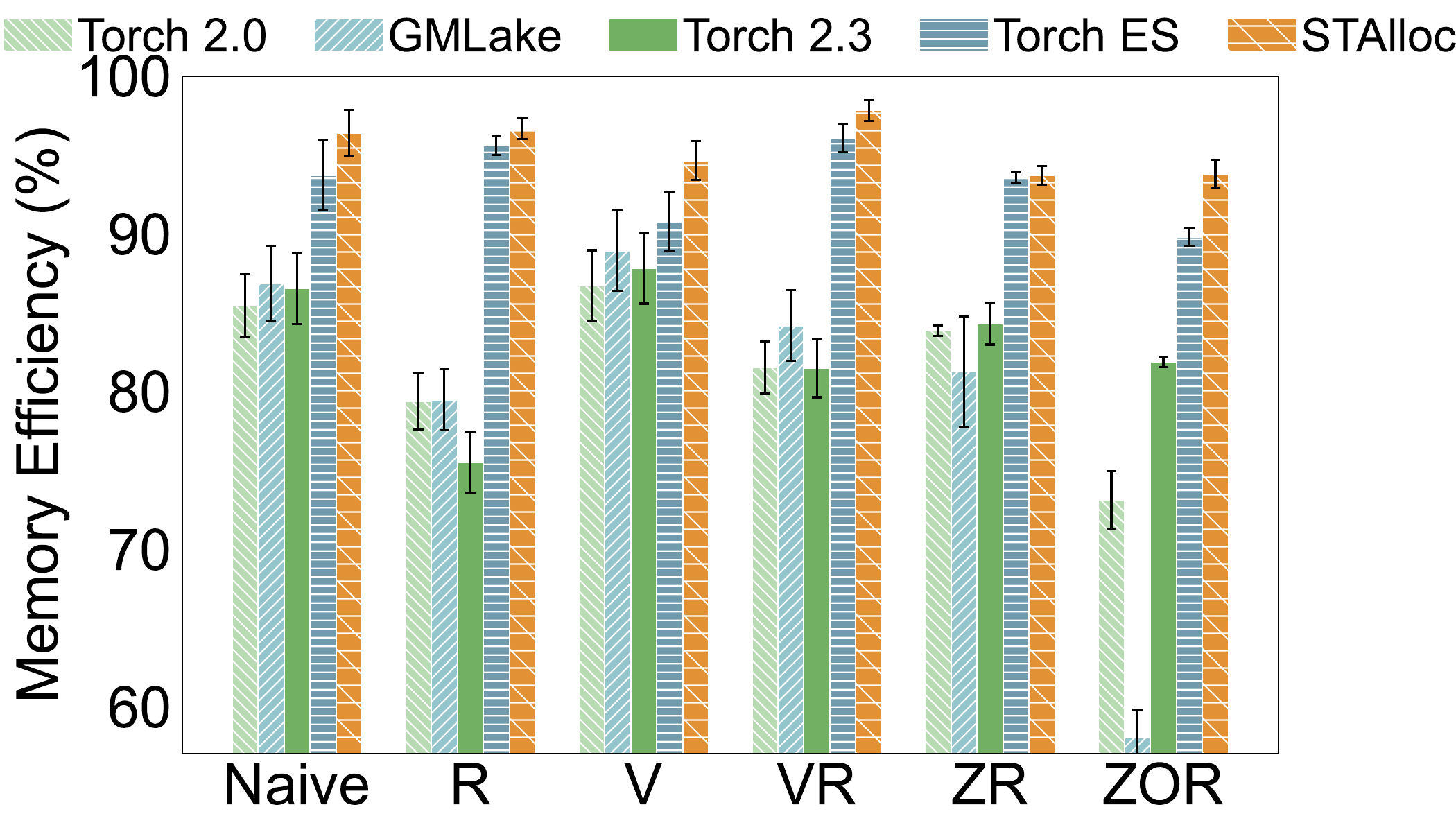}
        \vspace{-15pt}
        \caption{Qwen1.5-MoE-A2.7B.}
        \label{fig:a800_moe}
    \end{subfigure}
    \vspace{-10pt}
    \caption{Comparison of memory efficiency on the 3 models among different allocators, 
    using different combinations of optimizations, namely recomputation (R), Virtual Pipeline (V), ZeRO (Z), and offload (O).}
    \label{fig:memory_utilization}

\end{figure*}

\section{Evaluation}
\label{sec:evaluation}

To gain an in-depth understanding of \system,
    we focus on the following aspects in the evaluation.
\textbf{(1) Performance.}
We show that \system can reduce fragmentation memory by 85.1\% on average (up to 100\%), saving up to 56.3GB GPU memory across dense/sparse models trained with a variety of frameworks, configurations, and scales.
    % outperforming state-of-the-art solutions by \fixme{?}$\times$.
\textbf{(2) Overhead.}
We demonstrate that \system's impact on end-to-end training throughput is negligible in all cases,
    and our plan synthesizer can efficiently produce an allocation plan in minutes even under complex allocation requests.
\textbf{(3) Performance Breakdown.}
We study the individual performance of the static and dynamic allocators,
    and show their impacts on the final performance of \system.

% \vspace{-10pt}
\subsection{Experimental Setup}
\label{sec:eval_setup}

\noindent
\textbf{Testbed.}
\system is evaluated on both NVIDIA and AMD GPU platforms.
% Our testbed consists of both platforms.
One configuration consists of 1 node equipped with an Intel Xeon Platinum 8358 128-Core CPU and 8 NVIDIA A800-80GB GPUs, which is used to evaluate various training optimization setups.
The other has up to 16 nodes, each equipped with an Intel Xeon Platinum 8558 192-Core CPU and 8 NVIDIA H200-141GB GPUs, and is used for scalability evaluation.
The AMD GPU platform has 8 nodes, each of which is equipped with AMD EPYC 7K62 48-Core Processor and 8 AMD MI210-64GB GPUs.

\noindent
\textbf{Models.}
We evaluate \system on 7 representative large-scale dense and sparse Mixture-of-Expert (MoE) models.
For dense models, we choose GPT-2~\cite{openai-gpt2} and Llama2-7B~\cite{llama} for experiments on multiple training configurations.
We use four models of varying sizes (including 7B, 14B, 32B, 72B) from the Qwen2.5~\cite{qwen2.5} series to demonstrate the scalability of our approach with respect to both model size and cluster size.
For sparse models, we choose Qwen1.5-MoE-A2.7B~\cite{qwen1.5-moe}, a MoE model with 16 billion parameters to evaluate the efficiency of \system on both multiple configurations as well as scalability and extendability on AMD platform.
% This also allows us to directly demonstrate the effectiveness of \system in both static and dynamic models.

\noindent
\textbf{Training Setup.}
We evaluate \system with multiple training setups,
    in terms of training frameworks and training optimization techniques.
For training frameworks,
    we choose the popular Megatron-LM~\cite{megatron}, and Colossal-AI \cite{colossal-ai}.
For training optimizations,
    we choose the pipeline parallelism schedule of Pipedream-1F1B~\cite{pipedream-1f1b}, Virtual Pipeline~\cite{megatron2} as parallelism-based optimizations. 
For non-parallelism-based optimizations, we consider activation recomputation \cite{recompute1}, offloading \cite{zero-offload}, and distributed optimizer (ZeRO~\cite{zero}), which contains all kinds of memory optimizations~\cite{training-survey}.

\noindent
\textbf{Baselines.}
We compare \system with state-of-the-art baselines, including:
\begin{itemize}[label=•, left=0pt, labelsep=0.5em, itemsep=0em]
    \item \textbf{PyTorch~\cite{pytorch}.}
    PyTorch employs a caching memory allocator for GPU memory management. 
    It reduces the overhead of frequent native GPU API calls by reusing previously freed memory blocks, improving performance and memory efficiency.
    \item \textbf{PyTorch expandable\_segments (PyTorch ES)~\cite{pytorch-doc}.}
    The expandable\_segments allocator in PyTorch introduces support for virtual memory, allowing memory segments to grow dynamically as needed.
    This feature is only available in PyTorch versions 2.1 and above.
    \item \textbf{GMLake~\cite{gmlake}.}
    GMLake leverages virtual memory stitching to unify non-contiguous memory blocks into a single virtual space for defragmenetation.
    We deployed it using the official Docker image provided in its repository~\cite{glake_repo}, whose PyTorch version is 2.0.
\end{itemize}

\noindent
\textbf{Metrics.}
We evaluate the performance of \system using three key metrics. 
First, memory efficiency is the ratio of the max allocated memory to the max reserved memory as explained in \S\ref{sec:low-mem-eff}.
Building on this, the fragmentation ratio represents the proportion of reserved memory that is not actually utilized, which equals to
(1 - memory efficiency).
To measure the end-to-end throughput of training and evaluate the overhead of \system, 
    we choose FLOPS (floating point operations per second) as the throughput metric,
    which is calculated by training frameworks per training iteration.

% \vspace{-10pt}
\subsection{Memory Efficiency and Defragmentation}
% \vspace{-5pt}
\label{sec:utilization}

\begin{figure*}[t]
    \centering
    \begin{subfigure}[b]{0.3\textwidth}
        % \centering
        \includegraphics[width=\linewidth]{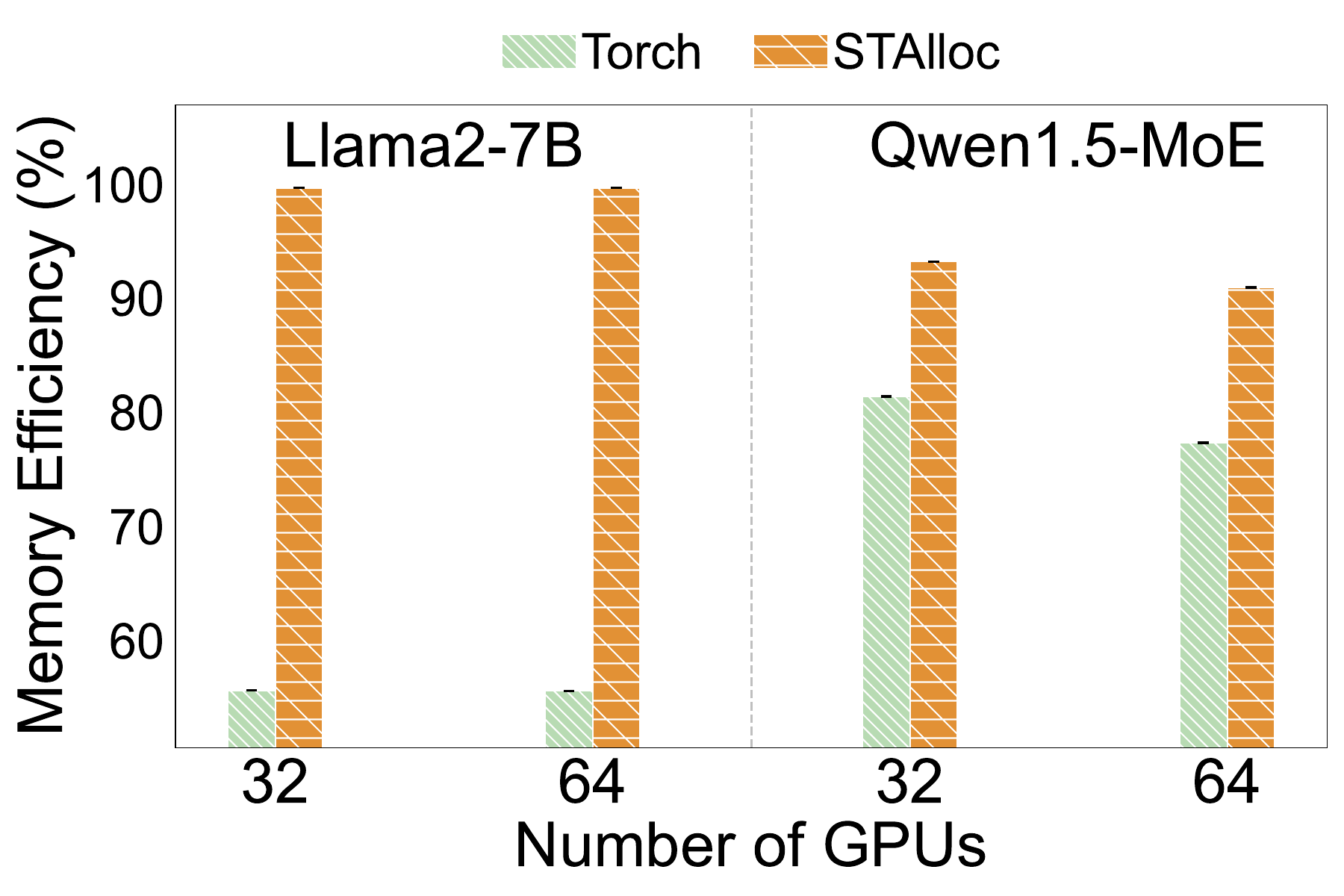}
        \vspace{-20pt}
        \caption{AMD GPU with recomputation.}
        \label{fig:amd_scale}
    \end{subfigure}
    \begin{subfigure}[b]{0.3\textwidth}
        % \centering
        \includegraphics[width=\linewidth]{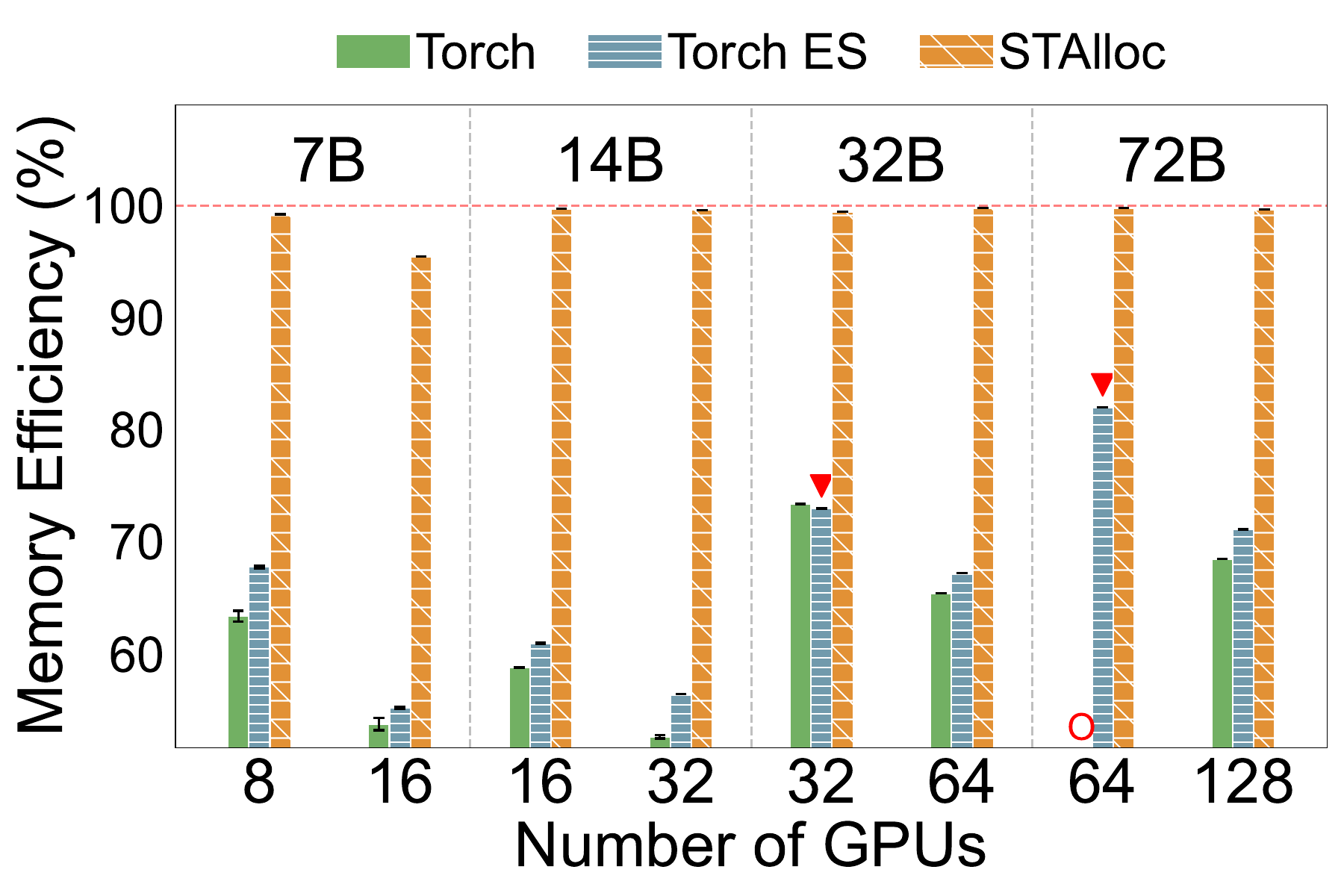}
        \vspace{-20pt}
        \caption{H200 GPU with recomputation.}
        \label{fig:h200_rcp}
    \end{subfigure}
    \begin{subfigure}[b]{0.3\textwidth}
        % \centering
        \includegraphics[width=\linewidth]{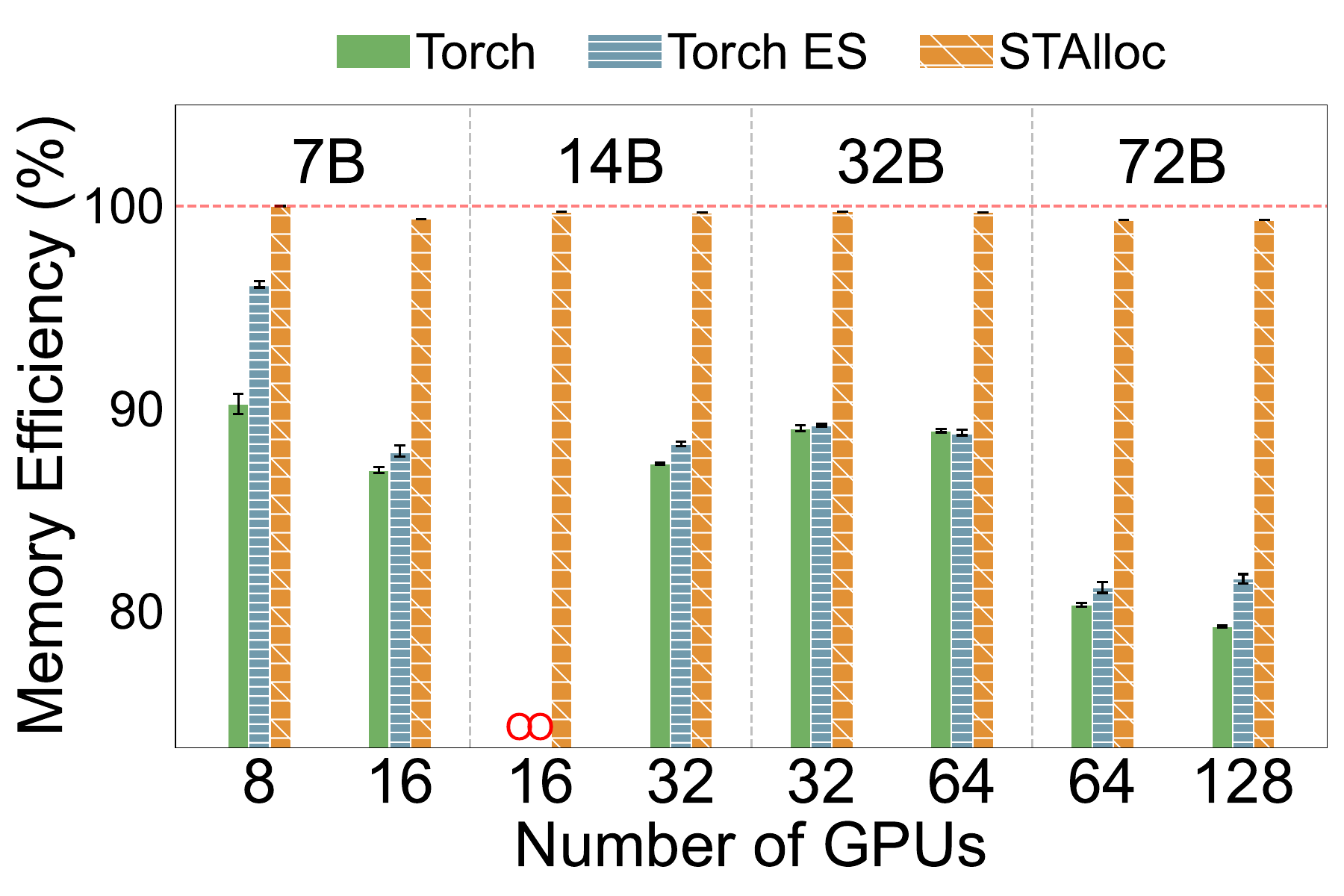}
        \vspace{-20pt}
        \caption{H200 GPU with virtual pipeline.}
        \label{fig:h200_vpp}
    \end{subfigure}
  \vspace{-10pt}
  \caption{Comparison of memory efficiency on different cluster scales and model sizes using optimization of recomputation or virtual pipeline. The red ``O'' means the case occurs OOM error, the red triangle means distinct throughput decrease.}
  \label{fig:main}
\end{figure*}

\begin{figure}[t]
    \centering
    \includegraphics[width=0.8\linewidth]{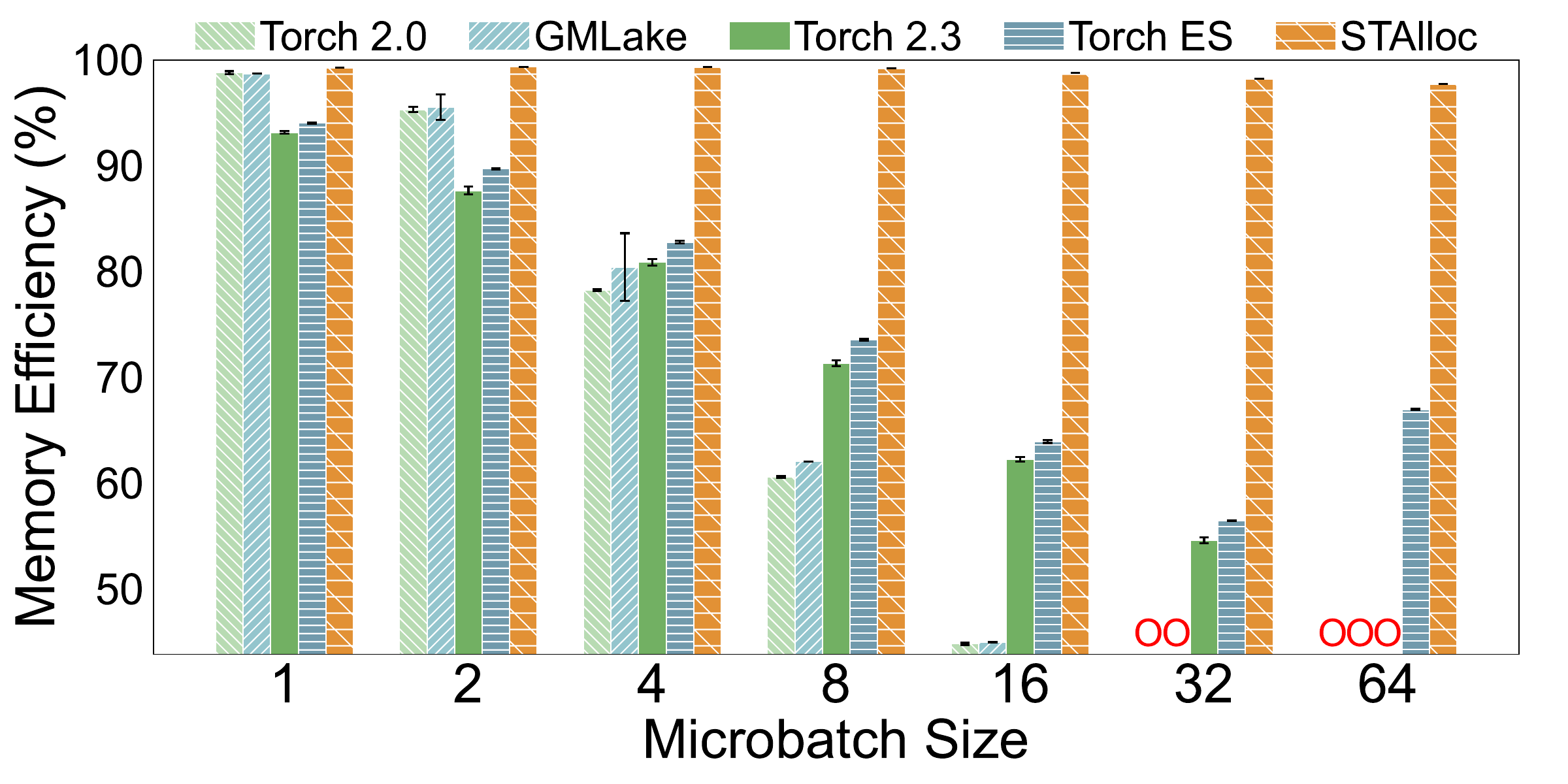}
    \vspace{-10pt}
    \caption{Memory efficiency under different micro-batch sizes when training Llama2-7B with recomputation.}
    \vspace{-10pt}
    \label{fig:mbs}
\end{figure}

\begin{figure*}[t]
    \begin{minipage}[t]{0.3\textwidth}
        \centering
        \includegraphics[width=\linewidth]{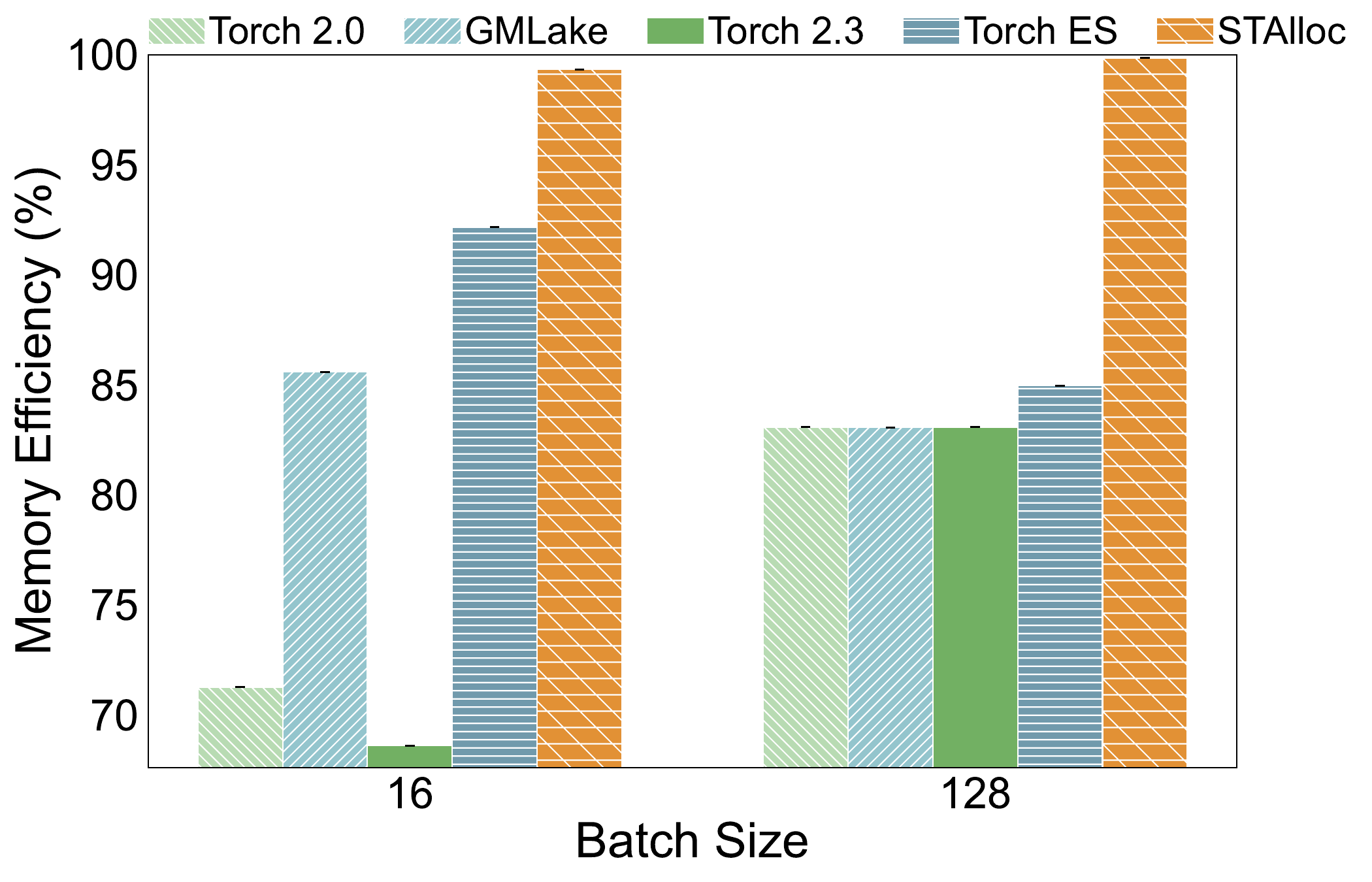}
        \vspace{-20pt}
        \caption{Memory efficiency comparison on Colossal-AI.}
        \label{fig:cai}
    \end{minipage}
    \hfill
    \begin{minipage}[t]{0.3\textwidth}
        \centering
        \includegraphics[width=\linewidth]{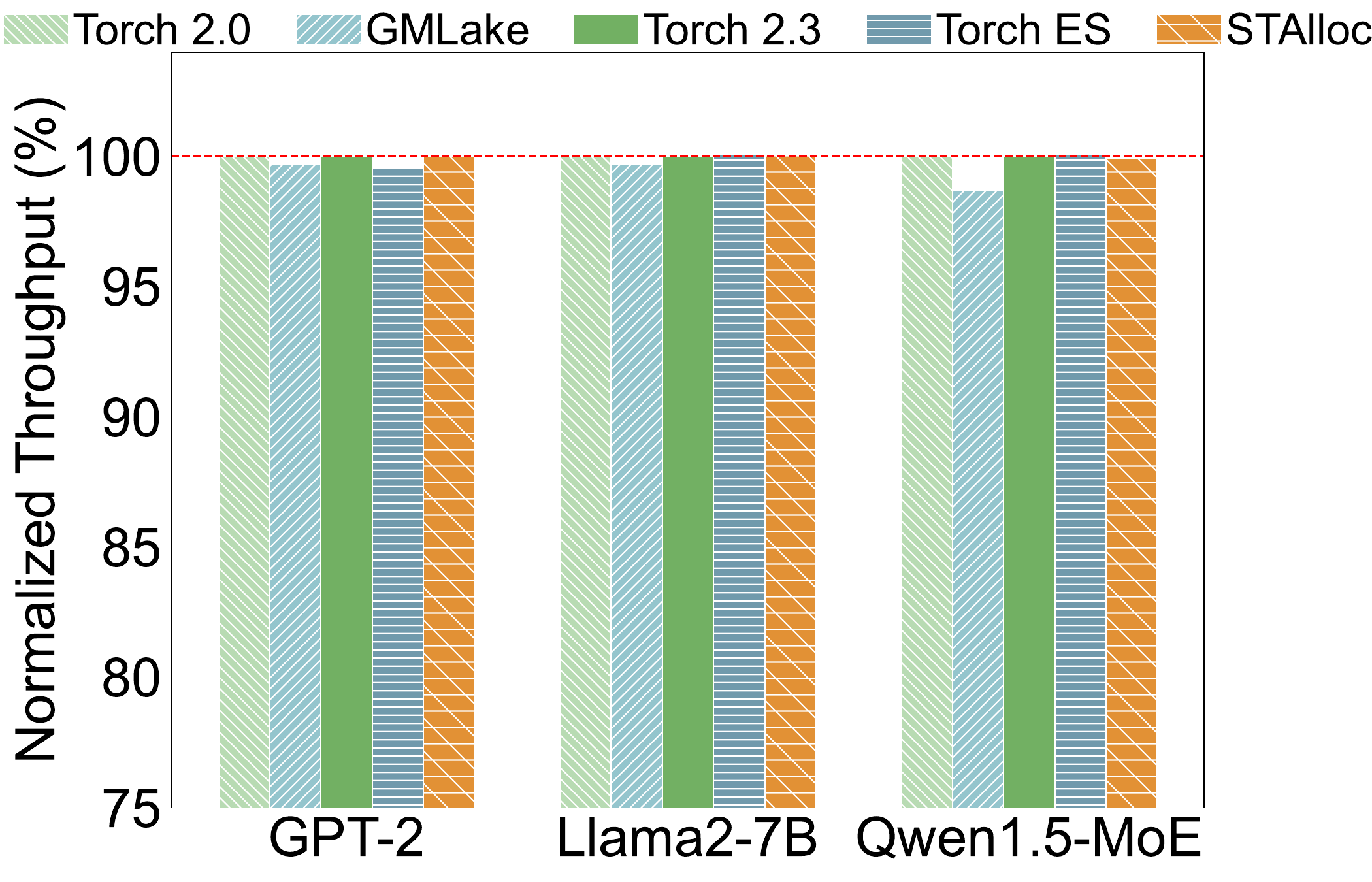}
        \vspace{-20pt}
        \caption{Training throughput comparison using different allocators.}
        \label{fig:throughput}
    \end{minipage}
    \hfill
        \begin{minipage}[t]{0.33\textwidth}
        \centering
        \includegraphics[width=\linewidth]{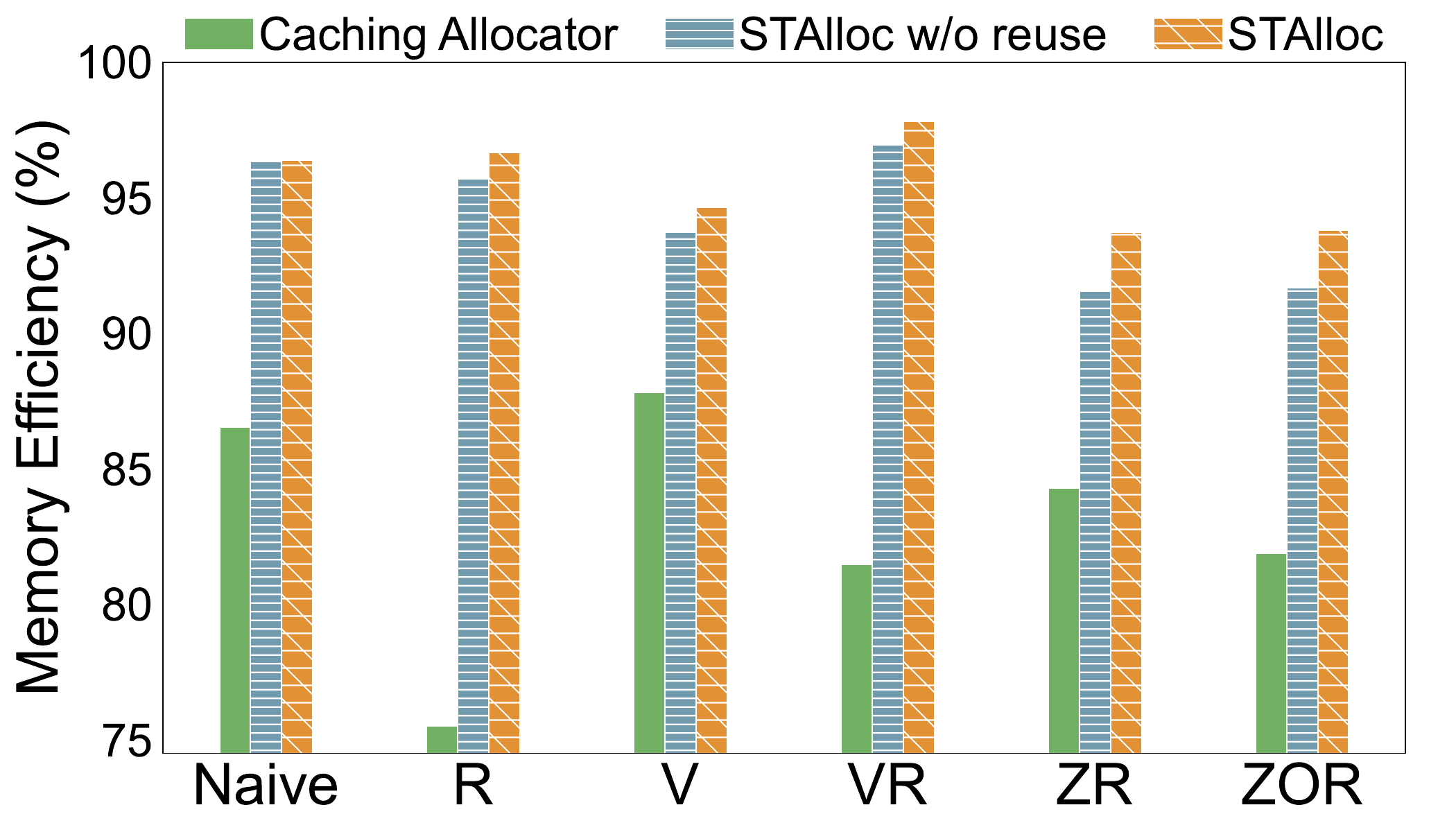}
        \vspace{-20pt}
        \caption{Performance breakdown under MoE model.}
        \label{fig:breakdown}
    \end{minipage}
    \hfill
    \vspace{-10pt}

\end{figure*}

\noindent
\textbf{Models and Optimization Techniques.}
For the model and optimization combination test, 
    the micro-batch sizes are set to the maximum feasible size that will not cause OOM following common practices~\cite{megatron2},
    i.e., 128, 4, and 8 for GPT-2, Llama2-7B, and Qwen1.5-MoE-A2.7B, respectively.

Our experiments used Megatron-LM as the training framework.
% Megatron-LM is employed for configurations with pipeline parallelism ($PP$); for dense models, we set tensor parallelism ($TP$) to 4 and $PP$ to 2, while MoE models used $TP=4$, $PP=2$, and expert parallelism $EP=4$ to hold the model. 
% In contrast, Megatron-DeepSpeed is used to evaluate techniques of ZeRO and ZeRO Offload. 
% Due to architectural incompatibility of the Qwen1.5-MoE-A2.7B model with Megatron-DeepSpeed for ZeRO-like evaluations, we instead use Megatron-LM's distributed optimizer and optimizer offload features as a substitute. 
% Finally, since evaluating ZeRO and ZeRO Offload requires $DP>1$, the settings on GPT-2 and Qwen1.5-MoE-A2.7B model exceed the eight A800 GPU memory capacity, we tested ZeRO with recomputation as an alternative for these cases.
Figure~\ref{fig:memory_utilization} shows the comparison of memory efficiency.
We can observe that for dense models that do not have dynamic layers,
    \system can achieve $>$95\% (up to 100\%) memory efficiency (i.e., fragmentation ratio < 5\%) in all cases,
    demonstrating the effectiveness of our spatio-temporal planning mechanism.
In comparison, PyTorch 2.3 produces 57.1\% to 90.6\% memory efficiency, GMLake produces 45.1\% to 88.1\% memory efficiency,
    and PyTorch ES yields 62.4\% to 93.2\% memory efficiency.
Compared to the baselines, \system reduces fragmentation memory by 90.3\%, 93.4\%, and 87.8\% on average,
    up to 100\%, reducing reserved memory up to 14.4GB (i.e., 18\% of GPU memory).
The most significant fragmentation reduction appears in the case of GPT-2 with ZeRO and recomputation,
    which is because the weight size of GPT-2 is relatively small compared to the other 2 models,
    and thus the proportion of activation tensors (whose lifecycle is affected by recomputation) among all the tensors is considerably larger than that of Llama2-7B.

For the MoE model with dynamic layers,
    \system still shows 93.7\% to 97.8\% memory efficiency in the evaluated cases, reducing the fragmentation ratio to 4.3\% on average.
Compared to PyTorch, GMLake, and PyTorch ES, whose fragmentation ratios are 17.7\%, 20.3\%, and 6.9\%, respectively.
\system occurs less fragmentation memory of 74.9\%, 77.2\%, and 34.0\%, respectively.    
In the MoE test, tuning the default GMLake defragmentation threshold (\texttt{fragLimit}) from 512~MB to 64~MB increased memory efficiency to 97.73\% but reduced training performance by 56.4\% over 50 iterations. The 64~MB threshold caused unstable virtual memory pools under MoE’s dynamic allocations, leading to frequent virtual memory operations (up to 1500 times per iteration, each taking around 30ms). A 512~MB threshold optimally balances memory efficiency and training performance.

\noindent
\textbf{Training Scales.}
We demonstrate the scalability of \system on the two different GPU platforms.
On the AMD platform, we train the Llama2-7B and Qwen1.5-MoE-A2.7B models on 4 nodes (32 GPUs) and 8 nodes (64 GPUs), respectively.
We excluded GMLake and PyTorch ES from this study, as GMLake does not support AMD GPUs, and the features of PyTorch ES are unavailable in our platform's PyTorch version (2.0).
All the training experiments are conducted with recomputation.
As shown in Figure~\ref{fig:amd_scale}, \system scales well for both the dense and MoE models.
The memory efficiency on both models achieves over 90\%, and up to 99.7\%.
In contrast, the PyTorch caching allocator exhibits memory efficiency below 60\% across all scales of the Llama2-7B model. 
This result shows that \system can reduce fragmentation memory of 22.8~GB, which is 35.6\% of GPU memory.
Moreover, for MoE models, when the cluster size increases from 32 to 64 GPUs, the memory efficiency drops below 80\%.

To further investigate the scalability of \system's memory efficiency as model and cluster sizes are concurrently augmented, we use four models of varying sizes from the Qwen2.5 series, including 7B, 14B, 32B, and 72B, on 8 to 128 NVIDIA H200 GPUs. 
The training configurations are either recomputation, as a memory optimization technique, or virtual pipeline, as a parallelism optimization strategy, thereby demonstrating \system's scalability across diverse scenarios.
GMLake is not included since it does not support PyTorch 2.6 on the current platform.

Under recomputation settings (Figure~\ref{fig:h200_rcp}), \system achieves 99.1\% memory efficiency, reducing fragmentation by over 98.5\% and 98.4\% compared to PyTorch 2.6 and PyTorch ES, respectively, saving 37.9~GB GPU memory on average, up to 56.3~GB. PyTorch ES showed throughput degradation: 15.0\% lower than PyTorch for the 32B model on 32 GPUs, while \system's throughput matched PyTorch within 0.02\%. For the 72B model on 64 GPUs, PyTorch faced OOM errors due to fragmentation, and PyTorch ES was 20.1\% slower than \system. PyTorch ES's overhead stems from frequent virtual memory API calls, whereas \system maintains high efficiency with minimal runtime penalties.

Under virtual pipeline settings as shown in Figure~\ref{fig:h200_vpp}, \system achieves memory efficiency over 99\% in all cases; 
    reduce fragmentation memory over 97.6\% and 97.4\% compared to PyTorch 2.6 and PyTorch ES, respectively, saving GPU memory of 15.7~GB on average.
We find that with the scaling of model and cluster sizes, the memory efficiency of PyTorch and PyTorch ES declined by 10.9\% and 15.0\%, respectively, while \system differs within 0.7\%.

When training the 14B model on 16 GPUs, only \system successfully completes the training without out-of-memory (OOM) error by reducing fragmentation.
To avoid OOM, PyTorch and PyTorch ES require disabling virtual pipeline, increasing the tensor parallelism degree, or introducing recomputation. 
The original training configuration outperforms these adjustments in training throughput by 5.4\% to 32.5\%, as shown in Table~\ref{tab:vpp_case}.
\emph{This indicates that by reducing fragmentation, \system enables more efficient training configurations and yields performance improvements.}
\begin{table}[t]
    \centering
    \small
    \caption{Train Qwen2.5-14B with 16 GPUs using different configurations. The original uses only VPP with $TP=2$.}
    \vspace{-10pt}
    \resizebox{\linewidth}{!}
{
    \begin{tabular}{c|c c c|c}
        \toprule
        Config & PyTorch & PyTorch ES & \system & \makecell{Throughput\\(TFLOPS)}\\
        \midrule
        Original & OOM & OOM & $\checkmark$ & 464.3 \\
        Disable VPP & OOM & $\checkmark$ & $\checkmark$ & 440.6 \\
        Recomputation & $\checkmark$ & $\checkmark$ & $\checkmark$ & 350.4 \\
        $TP = 4$ & $\checkmark$ & $\checkmark$ & $\checkmark$ & 431.5 \\
        \bottomrule
    \end{tabular}}
    \vspace{-15pt}
    \label{tab:vpp_case}
\end{table}

\noindent
\textbf{Micro-Batch Sizes.}
Given that activation memory usage during training is directly proportional to microbatch size, and that larger microbatch sizes typically enhance operator computational efficiency~\cite{megatron2}, we conducted further experiments across a range of microbatch sizes.
We conduct the experiments for micro-batch sizes 1, 2, 4, 8, 16, 32, and 64, 
    training Llama2-7B with recomputation on Megatron-LM.
As shown in Figure~\ref{fig:mbs}, \system yields the best and similar (around 99\%) memory efficiency regardless of the micro-batch size,
    while the other allocators generally performs worse as the micro-batch size increases,
    mostly because the increasing size of the activation tensors affected by recomputation.
This proves \system's robustness against memory-related training configurations in practice.

\noindent
\textbf{Training Frameworks.}
To evaluate \system's generalizability across high-level training frameworks,
    we also apply \system to Colossal-AI~\cite{colossal-ai},
    another representative training framework shipped with a variety of memory optimizations.
We train GPT-2 on Colossal-AI with tensor offload and ZeRO-3~\cite{zero} with two different batch sizes.
As depicted in Figure~\ref{fig:cai}, \system still performs better than the other allocators,
    demonstrating \system's general applicability across training frameworks.
% An interesting anecdote of using \system on Colossal-AI is that despite training a static dense model,
%     the framework still generates a small amount of dynamic tensors (an average of 3 tensors of 128~KB per iteration) in the process.
% This triggers \system's safety measure (cf. \S\ref{sec:implementation}) in the runtime allocator,
%     enabling the successful execution of the training run.
% \vspace{-10pt}
\subsection{Overhead Analysis}
\label{sec:eval_overhead}
We next evaluate \system's potential impact on the end-to-end training throughput,
    as well as the efficiency of the allocation profiler and plan synthesizer facing different numbers of allocation requests.

\noindent
\textbf{Overhead of Allocators in Training Throughput.}
Figure~\ref{fig:throughput} shows the normalized end-to-end training throughput when training the 3 test models on Megatron-LM using different allocators.
Specifically, GMLake is normalized against PyTorch 2.0,
    while PyTorch ES and \system are normalized against PyTorch 2.3 for fairness.
All the experiment settings adopt recomputation.
We can see that none of the allocators incur noticeable throughput degradation.
In particular, \system's throughput difference with the vanilla PyTorch 2.3 is $<$0.05\% in all cases,
    which are most likely due to hardware performance fluctuation.
It is worth noting that virtual memory–based GPU memory allocation methods have shown significant drops in training throughput under specific scenarios as discussed in \S\ref{sec:utilization}. 
GMLake exhibits such behavior in MoE models, and PyTorch ES demonstrates it in recomputation-heavy settings.
While these approaches help reduce memory fragmentation, the runtime overhead introduced by virtual memory operations can become non-negligible, ultimately impacting training performance.
% Here we show the throughput results using the balanced threshold 512~MB,
%     while the threshold with high memory efficiency leads to as low as 43.6\% normalized throughput.

The above throughput comparison uses identical training configurations. 
Thanks to \system's ability to reduce GPU memory usage without incurring extra runtime overhead, it enables the use of more memory-intensive configurations without triggering out-of-memory (OOM) errors. 
As a result, \system can achieve higher training throughput.

\noindent
\textbf{Profiling and Plan Synthesis Time.}
\begin{table}[t]
    \centering
    \caption{Profile and plan synthesis time in different training configuration. $Num$ is the number of requests within one iteration. -N and -R represent the configuration without/with recomputation, respectively.}
    \small
    \vspace{-10pt}
    \begin{tabular}{c|c|c|c}
    \toprule
        Config & $Num$ & $T_{profile} (s)$ & $T_{plan}(s)$ \\
        \midrule
        GPT-2-N & 12785 & 78.82 & 24.36 \\
        GPT-2-R & 16569 & 100.19 & 21.93 \\
        Llama2-7B-N & 66529 & 204.73 & 104.96 \\
        Llama2-7B-R & 86721 & 278.41 & 136.34 \\
        Qwen1.5-MoE-N & 196759 & 273.74 & 374.18 \\
        Qwen1.5-MoE-R & 281669 & 362.20 & 145.40 \\
        \bottomrule
    \end{tabular}
    \label{tab:plan_overhead}
    \vspace{-15pt}
\end{table}
To understand the efficiency of ahead-of-time planning,
    we further delve into the profile and plan synthesis time for different settings with varying complexity in terms of the number of total allocation requests that need to be planned per training iteration.
As shown in Table~\ref{tab:plan_overhead},
    the Allocation Profiler utilizing CUDA malloc/free, requires a runtime for minutes for three iterations, approximately 10\% to 30\% of the speed using PyTorch caching allocator. 
Given that profiling requires only three iterations, this overhead is deemed acceptable.
The plan synthesis time is around 2 minutes, up to 6 minutes for complex cases,
    and only around 20 seconds for simpler cases.
In the case of MoE models, the plan synthesis time in the configuration without recomputation markedly exceeds that of the configuration with recomputation. 
This disparity occurs because recomputation leads to the immediate deallocation of activation tensors within the same dynamic layer following their forward pass allocation. 
Conversely, in the absence of recomputation, these activation tensors must be preserved from their forward pass allocation until the corresponding dynamic layer in the backward pass to be freed. 
Consequently, during the plan generation phase, the configuration without recomputation results in a larger number of {\em HomoLayer Groups} when classifying dynamic requests. 
This, in turn, increases the quantity of associated {\em Dynamic Reuse Space} that needs to be interrogated, thereby prolonging the plan synthesis time.
% In comparison, a brute-force solution involves computation complexity at the magnitude of $10^{450K}$,
%     which is impossible to use.

% \vspace{-5pt}
\subsection{Performance Breakdown}
To understand the performance contribution of the static and dynamic allocators in \system, we evaluate the performance breakdown of \system when training the Qwen1.5-MoE-A2.7B model with the same setting in \S\ref{sec:utilization}.
To this end, we sequentially disable the dynamic allocator reusing {\em Static Allocation Plan} (mentioned as \system w/o reuse), and the static allocator (mentioned as Caching Allocator, which is the vanilla PyTorch caching Allocator), and measure the corresponding memory efficiency in the above cases.

\noindent
\textbf{Static Allocator.}
The results in Figure~\ref{fig:breakdown} indicate that \system with only the Static Allocation Plan reduces fragmentation memory by 70.2\% compared to PyTorch Caching Allocator. 
This reduction in fragmentation memory accounts for 91\% of the total fragmentation memory reduction achieved by the complete \system system relative to PyTorch.
Static planning accounts for the predominant share of the defragmentation result, primarily because static memory allocations form a substantial majority (from 73.4\% to 99.3\%) of the total memory allocation, as shown in Table~\ref{tab:composition}.

\begin{table}[t]
    \centering
    \small
    \caption{Composition of allocation types.\vspace{-10pt}}
    \resizebox{\linewidth}{!}
{
    \begin{tabular}{c|c c c c c c}
        \toprule
        Allocation type & None & R & V & VR & ZR & ZOR \\
        \midrule
        Total (GB) & 59.51 & 32.36 & 62.78 & 33.07 & 44.65 & 44.70 \\\hline
        Static (GB) & 44.68 & 31.39 & 46.10 & 31.83 & 44.62 & 44.40 \\\hline
        \makecell[c]{Dynamic fallback \\w/o reuse (GB)} & 15.19 & 1.61 & 17.80 & 1.78 & 2.99 & 1.95 \\\hline
        \makecell[c]{Dynamic fallback \\with reuse (GB)} & 15.19 & 1.12 & 17.22 & 1.70 & 1.92 & 1.55 \\
        \bottomrule
    \end{tabular}
    }
    \label{tab:composition}
    \vspace{-10pt}
\end{table}
\noindent
\textbf{Dynamic Allocator.}
Compared with \system without dynamic reuse, the full \system reduces memory fragmentation by an additional 22.9\%, mainly by lowering fallback allocations to the caching allocator.
As shown in Table~\ref{tab:composition}, enabling dynamic reuse decreases the number of requests falling back to the caching allocator. This benefit is most evident under recomputation, where caching allocations drop by 24.9\%. Without recomputation, the impact is smaller.

The difference stems from how recomputation affects memory lifespans. Without recomputation, activation memory is allocated during the forward pass and held until the backward pass, causing dynamic and static allocation requests' lifespans to fully overlap. This results in a peak memory usage close to the sum of both.
With recomputation, activation memory is released immediately after the forward pass, so static and dynamic requests do not overlap in time. As a result, dynamic requests can reuse idle regions in the static pool, reducing overall peak usage.

% \vspace{-10pt}
\section{Related Work}

% This section reviews related researches in efficient dynamic allocators for GPU memory,
    % static allocators in machine learning (ML) compilers,
    % as well as generic memory defragmentation techniques in memory systems.

\textbf{Online GPU Allocators.}
To reduce memory fragmentation and improve allocation efficiency,
    a plethora of online GPU memory allocators~\cite{halloc,fast-dynamic-allocator,scatteralloc,register-alloc} have been developed.
Dynamic allocators operate atop the native GPU memory
APIs (e.g., \texttt{cudaMalloc}) in a similar manner as the caching allocator of PyTorch. Differently, such allocators are meant to run on GPU threads alongside GPGPU applications, rather than managing GPU memory from the host like PyTorch and GMLake.
To reduce fragmentation, they usually adopt sophisticated allocation policies such as the slab and buddy systems.
Also, to achieve high-throughput allocations on GPUs, 
    they have proposed scalable synchronization primitives across the massive threads of GPUs~\cite{gelado}.
As a pluggable allocator of PyTorch,
    \system also chooses to manage GPU memory from the host to improve usability and programmability.
    
\if 0
Existing memory allocators commonly used in model training include those integrated within deep learning frameworks~\cite{tensorflow} and their optimized versions.
PyTorch~\cite{pytorch} includes a caching allocator to manage GPU memory, which helps improve memory allocation efficiency by reusing previously allocated memory blocks to minimize overhead.
Building upon this, GMLake~\cite{gmlake} introduces a memory management mechanism based on CUDA virtual memory. 
By stitching fragmented memory blocks, this approach significantly improves memory utilization.

These methods manage and organize allocation requests entirely during runtime, offering greater versatility across various scenarios. 
However, runtime decision-making may lead to suboptimal placement strategies and introduce additional time overhead.
Compared to this, our work focuses on memory management ahead of runtime, which contains more thorough optimization based on the regularity of memory activities.
\fi

\noindent
\textbf{Generic Memory Defragmentation Techniques.}
Memory defragmentation has been studied and discussed in various scenarios~\cite{fast-multicore-defrag,memory-frag-solved,eddy} beyond GPU applications.
Previous work~\cite{parallel-dfrag-gpu, marlow-copy-based,sapphire-copy-based,siegwart-copy-based} has proposed defragmentation strategy based on data movement or copying.
These approaches are mainly deployed in the real-time system with unpredictable runtime behaviors, 
    which result in complex defragmentation strategies with high runtime overhead.

% \system instead focuses on the training of large-scale models, where allocations exhibit temporal and spacial regularity that can be leveraged to  achieve low fragmentation without additional data movements.
\noindent
\textbf{Machine Learning Compilers.}
% ML compilers that lower high-level computation graphs to GPU device instructions also need to manage the memory allocation of the graphs.
% Given that the static nature of the graphs,
%     compilers can generate memory placement before their execution by analyzing static control dependencies, 
%     enabling optimized memory layouts during runtime.
% Current ML compilers~\cite{telamalloc, minimalloc, xla} usually handle this by heuristics or solvers focusing on a specific problem.
% TVM~\cite{tvm}, TFLite~\cite{tensorflow_micro} propose algorithms using greedy heuristics to find a reasonable allocation placement.
% To fully explore the search space and obtain improved memory placement strategies, Checkmate~\cite{checkmate} employs solver for optimal rematerialization to refine the results.
% In contrast, \system handles more complex dynamic allocations by combining ahead-of-time planning with real-time allocation. 
Machine Learning compilers that convert high-level computation graphs to GPU instructions must manage memory allocation for these graphs. 
The compilers pre-analyze control dependencies to optimize memory layouts before execution. 
Compilers like TVM~\cite{tvm} and TFLite~\cite{tensorflow_micro} use greedy heuristics for reasonable allocation, while Checkmate~\cite{checkmate} employs solvers for optimal rematerialization to improve results.
Unlike deep ML compilers that organize memory allocation and deallocation at the computation graph level, \system manages memory requests at the level of the overall model execution. 
These approaches are complementary and orthogonal with our work.
% These methods rely entirely on pre-runtime compilation optimizations for memory placement strategies. 
% Due to the complexity of the optimization algorithms, these approaches are primarily targeted at small machine learning models.
% In contrast, \system does not impose restrictions on static data flow. 
% Instead, we complement the pre-runtime placement strategy generation with real-time allocation to handle memory activities with dynamic patterns.
% Additionally, \system targets large-scale model training, where the scale of memory activities is significantly higher than that of standard machine learning models. 
% This results in greater challenges in analysis and optimization.
% These methods depend solely on pre-runtime compilation for memory placement, making them suitable for small machine learning models due to the complexity of the algorithms. 

% \vspace{-10pt}
\section{Conclusion}
This work presents the design, implementation, and evaluation of \system,
    a novel memory allocation system that significantly improves the memory utilization of large-scale model training.
\system builds on our insight of spatio-temporal regularity in model training allocation requests to combine ahead-of-time memory layout planning with runtime profile-guided allocation.
Extensive evaluations show that \system significantly outperforms state-of-the-art solutions in terms of both effectiveness and efficiency.
Complementary to existing runtime defragmentation methods,
    we believe \system demonstrates the powerful potential of fusing proactive pre-runtime planning with reactive runtime decision-making.

\begin{acks}
    We thank the reviewers and our shepherd, Christoph Kirsch, for their insightful comments. 
    This work was supported by the National Key R\&D Program of China (2023YFB4502200), the National Natural Science Foundation of China (62325405, 62504139, U24B6015), Beijing Natural Science Foundation (QY24247, L242018, L257010), Beijing National Research Center for Information Science, Technology (BNRist), Beijing Innovation Center for Future Chips, and State Key laboratory of Space Network and Communications.
\end{acks}
\bibliographystyle{ACM-Reference-Format}
\bibliography{reference}

\appendix
\section{Artifact}
The artifact code is available at Zenodo: 

\noindent
https://zenodo.org/records/17173036
% \newpage
% \input{EuroSys26_ArtifactAppendix_template}

\end{document}